\documentclass[sigconf, nonacm]{acmart}

\newcommand\vldbdoi{XX.XX/XXX.XX}
\newcommand\vldbpages{XXX-XXX}
\newcommand\vldbvolume{14}
\newcommand\vldbissue{1}
\newcommand\vldbyear{2020}
\newcommand\vldbtitle{\shorttitle} 
\newcommand\vldbavailabilityurl{http://vldb.org/pvldb/format_vol14.html}

\usepackage{xspace}
\newcommand{\sys}{\textsc{Hyper-Tune}\xspace}

\usepackage{subfigure}
\usepackage{multirow}
\usepackage{bm}
\usepackage{bbm}
\usepackage{dsfont}
\usepackage{color, soul}
\usepackage{listings}
\usepackage{xcolor}
\usepackage{multibib}
\newcites{A}{References}

\usepackage[linesnumbered,vlined,ruled,noend]{algorithm2e}

\newcommand{\tabincell}[2]{\begin{tabular}{@{}#1@{}}#2\end{tabular}}
\newcommand{\para}[1]{{\vspace{2pt} \bf \noindent #1 \hspace{1pt}}}
\definecolor{codegray}{rgb}{0.5,0.5,0.5}

\usepackage{enumitem}

\usepackage{wrapfig}

\begin{document}
\title{\sys: Towards Efficient Hyper-parameter Tuning at Scale}
%\subtitle{Volcano Style Processing
%Meets AutoML [Scalable Data Science]}
\subtitle{[Scalable Data Science]}

\renewcommand{\shorttitle}{\sys: Towards Efficient Hyper-parameter Tuning at Scale}

%%
%% The "author" command and its associated commands are used to define the authors and their affiliations.

\author{Yang Li$^{\dagger\ddagger}$, Yu Shen$^{\dagger\ddagger}$, Huaijun Jiang$^{\dagger\ddagger}$, Wentao Zhang$^\dagger$, Jixiang Li$^\ddagger$, Ji Liu$^\ddagger$, Ce Zhang$^\mathsection$, Bin Cui$^\dagger$}
\affiliation{
$^\dagger$EECS, Peking University~~~~~$^\mathsection$System Group, ETH Z\"urich~~~~~$^\ddagger$AI Platform, Kuaishou Technology
}
\affiliation{
$^\dagger$\{liyang.cs, shenyu, jianghuaijun, wentao.zhang, bin.cui\}@pku.edu.cn~~~~~\\$^\ddagger$lijixiang@kuaishou.com~~ $^\ddagger$jiliu@kwai.com~~ $^\mathsection$ce.zhang@inf.ethz.ch
}

\renewcommand{\shortauthors}{Li et al.}

%%
%% The abstract is a short summary of the work to be presented in the
%% article.
\begin{abstract}
The ever-growing demand and complexity of machine learning are putting pressure on hyper-parameter tuning systems: \textit{while the evaluation cost of models continues to increase, the scalability of state-of-the-arts starts to become a crucial bottleneck}.
In this paper, inspired by our experience when deploying hyper-parameter tuning in a real-world application in production and the limitations of existing systems,
we propose \sys, an efficient and robust distributed hyper-parameter tuning framework. 
Compared with existing systems, \sys highlights multiple system optimizations, including (1) automatic resource allocation, (2) asynchronous scheduling, and (3) multi-fidelity optimizer.
We conduct extensive evaluations on benchmark datasets and a large-scale real-world dataset in production.
Empirically, with the aid of these optimizations, \sys outperforms competitive hyper-parameter tuning systems on a wide range of scenarios, including XGBoost, CNN, RNN, and some architectural hyper-parameters for neural networks.
Compared with the state-of-the-art BOHB and A-BOHB, \sys achieves up to $11.2\times$ and $5.1\times$ speedups, respectively.
\end{abstract}

\settopmatter{printfolios=true}
\maketitle

%%% do not modify the following VLDB block %%
%%% VLDB block start %%%
\begingroup\small\noindent\raggedright\textbf{PVLDB Reference Format:}\\
Yang Li, Yu Shen, Huaijun Jiang, Wentao Zhang, Jixiang Li, Ji Liu, Ce Zhang, and Bin Cui. \vldbtitle. PVLDB, \vldbvolume(\vldbissue): \vldbpages, \vldbyear.\\
\href{https://doi.org/\vldbdoi}{doi:\vldbdoi}
\endgroup
\begingroup
\renewcommand\thefootnote{}\footnote{\noindent
This work is licensed under the Creative Commons BY-NC-ND 4.0 International License. Visit \url{https://creativecommons.org/licenses/by-nc-nd/4.0/} to view a copy of this license. For any use beyond those covered by this license, obtain permission by emailing \href{mailto:info@vldb.org}{info@vldb.org}. Copyright is held by the owner/author(s). Publication rights licensed to the VLDB Endowment. \\
\raggedright Proceedings of the VLDB Endowment, Vol. \vldbvolume, No. \vldbissue\ %
ISSN 2150-8097. \\
\href{https://doi.org/\vldbdoi}{doi:\vldbdoi} \\
}\addtocounter{footnote}{-1}\endgroup
%%% VLDB block end %%%

%%% do not modify the following VLDB block %%
%%% VLDB block start %%%
\ifdefempty{\vldbavailabilityurl}{}{
\vspace{.3cm}
\begingroup\small\noindent\raggedright\textbf{PVLDB Availability Tag:}\\
The source code of this research paper has been made publicly available at
\textbf{\url{https://github.com/PKU-DAIR/HyperTune}}.
\endgroup
}
%%% VLDB block end %%%

%%%%%%%%%%%%%%%%%%%
%% INTRODUCTION
%%%%%%%%%%%%%%%%%%%
\section{Introduction}
\label{sec:intro}
Recently, researchers in the database community have been
working on integrating machine learning functionality into data management systems, e.g., 
SystemML~\cite{ghoting2011systemml}, SystemDS~\cite{boehm2020systemds}, Snorkel~\cite{ratner2020snorkel}, ZeroER~\cite{wu2020zeroer}, TFX~\cite{baylor2017tfx, breck2019data}, 
``Query 2.0''~\cite{wu2020complaint}, Krypton~\cite{nakandala2019incremental}, Cerebro~\cite{nakandala2020cerebro}, ModelDB~\cite{vartak2016modeldb}, 
MLFlow~\cite{zaharia2018accelerating}, 
HoloClean~\cite{rekatsinas2017holoclean}, 
and NorthStar~\cite{kraska2018northstar}.
AutoML systems~\cite{yao2018taking,automl_book,olson2019tpot}, an emerging type of data system, significantly raise the level of abstractions of building ML applications. 
While hyper-parameters drive both the efficiency and quality of machine learning applications, automatic hyper-parameter tuning attracts intensive interests from both practitioners and researchers~\cite{li2019qtune,zhang2021restune,falkner2018bohb,li2018massively,jaderberg2017population,liu2018darts,real2019regularized}, and becomes an indispensable component in many data systems~\cite{Li2021VolcanoMLSU,wang2018rafiki,shang2019democratizing}. 
% In this paper, we focus on a fundamental building block to enable efficient automatic hyper-parameter tuning.

% As machine learning (ML) models become prevalent in industrial-scale applications, automatic hyper-parameter tuning attracts intensive interests from both practitioners and researchers~\cite{falkner2018bohb,li2018massively,jaderberg2017population,liu2018darts,real2019regularized}. 
An efficient tuning system, which usually involves sampling and evaluating configurations iteratively, needs to support a diverse range of hyper-parameters, from learning rate, regularization, to those 
closely related to neural network architectures such as operation types, \# hidden units, etc.
Automatic tuning methods (e.g., Hyperband~\cite{li2018hyperband} and  BOHB~\cite{falkner2018bohb}) have been studied to tune a wide range of models, including XGBoost~\cite{chen2016xgboost}, recurrent neural networks~\cite{hochreiter1997long}, convolutional neural networks~\cite{he2016deep}, etc.
% As of today, they have become indispensable components in AutoML systems.
In this paper, we focus on building efficient and scalable tuning systems.

\iffalse
As machine learning (ML) models become prevalent in industrial-scale applications, automatic hyper-parameter tuning attracts intensive interests from both practitioners and researchers~\cite{falkner2018bohb,li2018massively,jaderberg2017population,liu2018darts,real2019regularized}. 
An efficient tuning system, which usually involves sampling and evaluating configurations iteratively, needs to 
support a diverse range of hyper-parameters, from 
learning rate, regularization, to those 
closely related to neural network architectures such as operation types, \# hidden units, etc.
Automatic tuning methods (e.g., Hyperband~\cite{li2018hyperband} and  BOHB~\cite{falkner2018bohb}) have been studied to tune a wide range of models, including XGBoost~\cite{chen2016xgboost}, recurrent neural networks~\cite{hochreiter1997long}, convolutional neural networks~\cite{he2016deep}, etc.
As of today, they have become indispensable components in modern AutoML systems~\cite{feurer2015efficient,automl_book,olson2019tpot}.
In this paper, we focus on a fundamental building block to enable efficient automatic hyper-parameter tuning.
\fi 

{\em \underline{Current Landscape}. }
Existing automatic hyper-parameter tuning methods include:
Bayesian optimization~\cite{hutter2011sequential,bergstra2011algorithms,snoek2012practical}, rule-based search, genetic algorithm~\cite{jaderberg2017population,olson2019tpot}, random search~\cite{bergstra2012random,dong2019bench}, etc.
Many of them have two flavors -- \textit{complete evaluation based search} and \textit{partial evaluation based search}.
To obtain the performance for each configuration, the complete evaluation based approaches~\cite{bergstra2011algorithms,hutter2011sequential,bergstra2012random} require complete evaluations that are usually computationally expensive. 
Instead, partial evaluation based methods~\cite{klein2016learning,klein2017fast,li2018hyperband,li2021mfes,falkner2018bohb} assign each configuration with incomplete training resources to obtain the evaluation result, thus saving the evaluation resources.

% XXX (talk about existing methods) \\

\iffalse
\begin{figure}[tb]
	\centering
		\scalebox{1.}[1.]{
			\includegraphics[width=1\linewidth]{images/intro_early_stopping.pdf}
		}
  \vspace{-2.5em}
  \caption{Resource assignment issue.}
  \vspace{-1.3em}
  \label{fig:early-stop}
\end{figure}
\fi

\begin{figure}[tb]
	\centering
		% Requires \usepackage{graphicx}
		\scalebox{0.95}[0.95]{
			\includegraphics[width=1\linewidth]{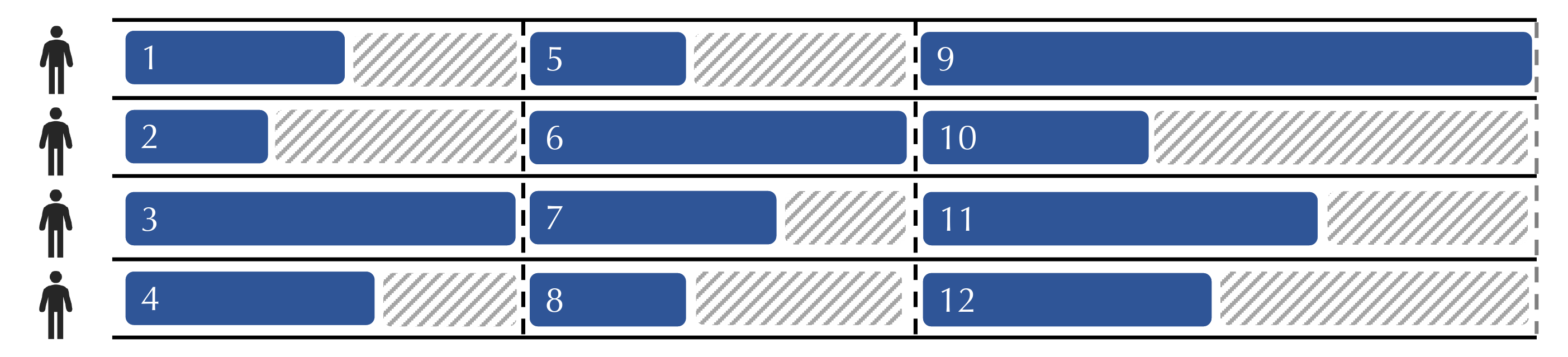}
	    }
	\vspace{-1em}
	\caption{Synchronous mechanism in hyper-parameter tuning, where each row represents a worker.
	The deep-blue areas correspond to the evaluation process of configurations; the striped areas refer to idle time.}
  	\label{fig:parallel-utilization}
\end{figure}

{\em \underline{An
Emerging Challenge in Scalability}.}
This paper is inspired by our efforts applying these latest methods to applications running at a large 
Internet company. 
One critical challenge arises from the \textbf{gap between the scalability of existing automatic tuning methods and the ever-growing complexity of industrial-scale models.
}
In recent years, we have witnessed that evaluating ML models are getting increasingly expensive as the size of datasets and models grows larger. 
For example, it takes days to train NASNet~\cite{zoph2018learning} to convergence on ImageNet, not to mention models like GPT-3~\cite{brown2020language} with hundreds of billions of parameters. 
Unfortunately, it is difficult for existing tuning methods to scale well to such tasks with ever-increasing evaluation costs, thus leading to a sub-optimal configuration for deployment.
When deploying existing approaches in large-scale applications, 
% we realize three limitations as follows:
we realize some limitations in the following three aspects:

{\bf (1) Design of Partial Evaluations.} Since the complete evaluation of a configuration is usually expensive (e.g., training deep learning models or training ML models on {\em large-scale} datasets), recent studies propose to evaluate configurations using partial resources (e.g., training models using a few epochs or a subset of the training set)~\cite{jamieson2016non,li2018hyperband,poloczek2017multi}.
However, \textit{how many training resources should be allocated to each partial evaluation?}
This question is non-trivial: 
(1) evaluations with small training resources could decrease evaluation cost, however, may be inaccurate to guide the search process; whereas (2) 
over-allocating resources could have the risk of high evaluation costs but diminishing returns from precision improvements.
\textit{How can we automatically decide on the right level of resource allocation
to balance the ``precision vs. cost'' trade-off in partial evaluations?}
This question remains open in state-of-the-arts~\cite{falkner2018bohb,abohb,li2021mfes}.

{\bf (2) Utilization of Parallel Resources.} Along with the rapid increase of evaluation cost, it comes with the rise of computation resources made available by industrial-scale clusters.
However, state-of-the-arts, such as BOHB~\cite{falkner2018bohb} and MFES-HB~\cite{li2021mfes}, often use a \textit{synchronous} architecture, which often cannot fully utilize all computation resources due to the synchronization barrier and are often sensitive to stragglers (See Figure~\ref{fig:parallel-utilization}).
ASHA~\cite{li2018massively} is able to remove these issues associated with synchronous promotions by incurring a number of inaccurate promotions, while this asynchronous promotion could hamper the sample efficiency when utilizing the parallel and distributed resources.
\textit{Thus, we need to explore an efficient asynchronous mechanism which pursues both sample efficiency and high utilization of parallel resources simultaneously}.

% In order to further improve these systems, we have to revisit the design of the synchronization mechanism.
% Not surprisingly, there are recent efforts in asynchronous communications~\cite{li2020system};
% however, as we will show later, 
% % \textit{we can further improve the asynchronous mechanism}.
% \textit{there still remains space towards efficient asynchronous mechanism}.

{\bf (3) Support of Advanced Multi-fidelity Optimizers.} 
While there are recent advancements in the design of Bayesian optimization methods, most, if not all, distributed tuning    systems~\cite{golovin2017google,falkner2018bohb,li2020system} have not fully utilized these advanced algorithms.
For example, while there are algorithms that can more effectively exploit the low-fidelity measurements generated by partial evaluations~\cite{li2021mfes}, many existing systems~\cite{feurer2015efficient,falkner2018bohb} still depend on vanilla Bayesian optimization methods that only use the high-fidelity measurements from the complete evaluations. 
\textit{Can we design a flexible system architecture to conveniently support drop-in replacement of different optimizers under the async/synchronous parallel settings?} This question is especially important from a system perspective.

\textbf{Contributions}.
% \para{Our Contributions. }
Inspired by our experience and observations deploying these state-of-the-art methods in our scenarios, in this paper, \textbf{(C.1)} we propose \sys, an efficient distributed automatic hyper-parameter tuning framework. 
\sys has three core components: \textit{resource allocator}, \textit{evaluation scheduler}, and  \textit{generic 
optimizer}, each of which corresponds to one
aforementioned question:
(1) To accommodate the first issue, we design a simple yet novel resource allocation method that could search for a good allocation via trial-and-error,
and this method can automatically balance the trade-off between the \textit{precision} and \textit{cost} of evaluations. 
(2) To mitigate the second issue, we propose an efficient asynchronous mechanism -- D-ASHA, a novel variant of ASHA~\cite{li2020system}. D-ASHA pursues the following two aspects simultaneously:
(i) \textit{synchronization efficiency}: the overhead of synchronization in wall-clock time, and 
(ii) \textit{sample efficiency}: the number of runs that the algorithm needs to converge.
(3) To tackle the third issue, we provide a modular design that allows us to plug in different hyper-parameter tuning optimizers.
This flexible design allows us to plug in MFES-HB~\cite{li2021mfes}, a recently proposed multi-fidelity optimizer. 
In addition, we also adopt an algorithm-agnostic sampling framework, which enables each optimizer algorithm to adapt to the sync/asynchronous parallel scenarios easily.
\textbf{(C.2)} 
We conduct extensive empirical evaluations on both publicly available benchmark datasets and a large-scale real-world dataset in production. 
\sys achieves strong anytime and converged performance and outperforms state-of-the-art methods/systems on a wide range of hyper-parameter tuning scenarios:
(1) XGBoost with nine hyper-parameters,
(2) ResNet with six hyper-parameters,
(3) LSTM with nine hyper-parameters,
and (4) neural architectures with six hyper-parameters.
Compared with the state-of-the-art BOHB~\cite{falkner2018bohb} and A-BOHB~\cite{abohb}, \sys achieves up to $11.2\times$ and $5.1\times$ speedups, respectively. 
In addition, it improves the AUC by 0.87\% in an industrial recommendation application with a billion instances.

\vspace{0.3em}
\textbf{Overview.}
This paper is organized as follows. We discuss the related work in Section~\ref{sec:related_work}.
In Section~\ref{sec:preliminary}, we review Bayesian optimization, synchronous Hyperband, as well as asynchronous ASHA. 
The proposed framework is presented in Section~\ref{sec:proposed_framework}. 
We provide empirical evaluations for hyper-parameter tuning problems in Section~\ref{sec:exp} and end this with the conclusion and future work in Section~\ref{sec:conclusion}.

\iffalse
As we will discuss in Section 2, while there has been a lot of recent progress in the field of hyper-parameter optimization, the key contribution of this paper is therefore to combine the strengths of several methods and build a practical and efficient system that fulfills all of these desiderata. 
Concretely, to address the first issue, we propose a methods to automatically design proper partial evaluations. In this way, the proposed system could maximize the strength of early-stopping, while keeping high efficiency.
Then we also introduce a new asynchronous paradigm to parallelize configuration evaluations in a distributed environment, which can make an effective use of parallel computing resources (2nd issue).
Furthermore, this system proposes a multi-fidelity surrogate framework (3rd issue), which can utilize the multi-fidelity measurements from both partial and complete evaluations to speed up the convergence of configuration search.
Our extensive empirical evaluations demonstrate that the proposed system could outperform the considered state-of-the-art methods or frameworks on a wide range of hyper-parameter tuning task types, including XGBoost, convolutional neural networks, recurrent neural networks, and neural architecture search.
\fi

\section{Related Work}
\label{sec:related_work}

Bayesian optimization (BO) has been successfully applied to hyper-parameter tuning~\cite{snoek2012practical,bergstra2011algorithms,hutter2011sequential,hutter2019automated,yao2018taking}.
% The main idea of BO is to use a probabilistic surrogate to model the relationship between a configuration $\bm{x}$ and its performance $f(\bm{x})$ (e.g., the validation error), and then use the surrogate to guide configuration sampling. 
% Since the evaluation of $f(\bm{x})$ for configuration $\bm{x}$ is getting more and more computationally expensive, many methods propose to evaluate configurations with partial evaluations, instead of the complete evaluations in BO. 
Instead of using complete evaluations, Hyperband~\cite{li2018hyperband} (HB) dynamically allocates resources to a set of random configurations, and uses the successive halving algorithm~\cite{jamieson2016non} to stop badly-performing configurations in advance.
BOHB~\cite{falkner2018bohb} improves HB by replacing random sampling with BO.
Two methods~\cite{domhan2015speeding,klein2016learning} propose to guide early-stopping via learning curve extrapolation.
Vizier~\cite{golovin2017google}, Ray Tune~\cite{liaw2018tune} and OpenBox~\cite{Li2021OpenBoxAG} also include a median stopping rule to stop the evaluations early.
In addition, multi-fidelity methods~\cite{klein2017fast,baker2017practical,hu2019multi,dai2019bayesian,li2021mfes,abohb} also exploit the low-fidelity measurements from partial evaluations to guide the search for the optimum of objective function $f$.
MFES-HB~\cite{li2021mfes} combines HB with multi-fidelity surrogate based BO.

% With the rise of parallel computing, leveraging distributed computational resources becomes a solution to deal with the ever-growing evaluation cost problem.
Many methods~\cite{gonzalez2016batch,azimi2010batch,kandasamy2017asynchronous} can evaluate several configurations in parallel instead of sequentially. 
However, most of them~\cite{gonzalez2016batch}, including BOHB~\cite{falkner2018bohb}, focus on designing batches of configurations to evaluate at once, and few support asynchronous scheduling.
% BOHB~\cite{falkner2018bohb} combines Hyperband with BO, where the evaluations at the same resource level are parallelized synchronously.
ASHA~\cite{li2018massively} introduces an asynchronous evaluation paradigm based on successive halving algorithm~\cite{jamieson2016non}.
In addition, Many approaches~\cite{kandasamy2018parallelised,alvi2019asynchronous} with asynchronous parallelization cannot exploit multiple fidelities of the objective; A-BOHB~\cite{abohb} supports asynchronous multi-fidelity hyper-parameter tuning. 
Searching architecture hyper-parameters for neural networks is a popular tuning application.
% , where it includes a discrete and graph-like search space and the target is to find the architecture with the best evaluation performance. 
Recent empirical studies~\cite{dong2019bench,siems2020bench} show that sequential Bayesian optimization methods~\cite{ma2019deep,white2019bananas,kandasamy2018neural,ru2020neural} achieve competitive performance among a number of NAS methods~\cite{real2019regularized,liu2018darts,awad2020differential,xu2019pc,zoph2018learning}, which highlights the essence of parallelizing these BO related methods.

A-BOHB~\cite{abohb} is the most related method compared with \sys, while it suffers from the first issue. % In addition, the design principles are quite different when dealing with the 2nd and 3rd challenges. 
BOHB~\cite{falkner2018bohb} lacks design to tackle the aforementioned three problems, and 
MFES-HB~\cite{li2021mfes} also faces these first and second issues. 
Instead, \sys is designed to accommodate the three issues simultaneously.

\label{sec:sh_hb}
\begin{figure}[t]
	\centering
		\scalebox{0.8}[0.8] {
		\includegraphics[width=1\linewidth]{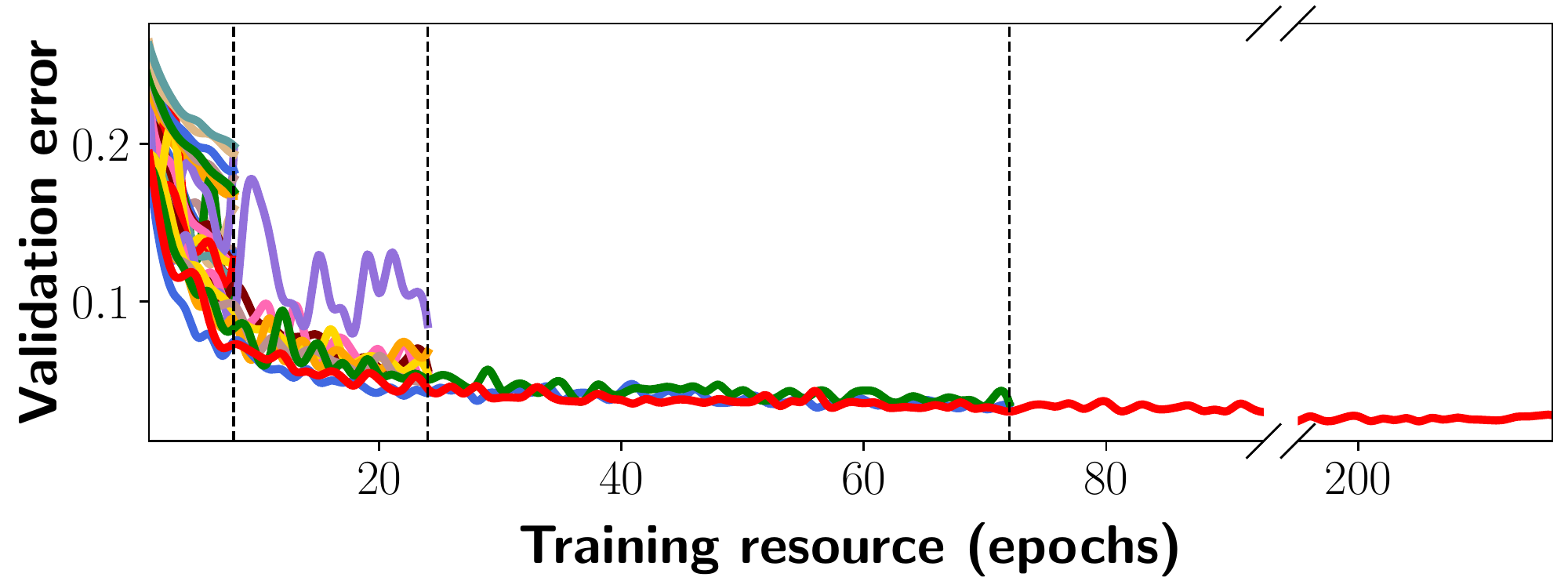}
         }
    \vspace{-1.2em}
	\caption{One iteration of successive halving algorithm (SHA) when tuning a CNN on MNIST, where $n_1=27$, $r_1=1$, $R=27$, $\eta=3$, and one unit of resource corresponds to 8 epochs.
	First, 27 configurations are evaluated with 1 unit of resource, i.e., 8 epochs ($n_1=27$ and $r_1=1$). 
	Then the top $\eta^{-1}$ configurations continue their evaluations with $\eta$ times units of resources (i.e., $n_2=27*\eta^{-1}=9$ and $r_2=r_1*\eta=3$). Finally, only one configuration is evaluated with the maximum  resource $R$.
	}
	\vspace{1.em}
    \label{fig:hb_sh}
\end{figure}

\section{Preliminary}
\label{sec:preliminary}

We define the hyper-parameter tuning as a black-box optimization problem, where the objective value $f(\bm{x})$ (e.g., validation error) reflects the performance of an ML algorithm with given hyper-parameter configuration $\bm{x} \in \mathcal{X}$. 
% , i.e., the objective value $f(\bm{x})$ (e.g., validation error), which reflects the performance of an ML algorithm with given hyper-parameter configuration $\bm{x} \in \mathcal{X}$, can only be obtained by evaluation.
The goal is to find the optimal configuration that minimizes the objective function $\bm{x}^{\ast}=\arg\min_{\bm{x} \in \mathcal{X}}f(\bm{x})$, and the only mode of interaction with $f$ is to evaluate the given configuration $\bm{x}$. 
In the following, we introduce existing methods for solving this black-box optimization problem, and these methods are the basic ingredients in \sys.

\vspace{-0.5em}
\subsection{Bayesian Optimization}
The main idea of Bayesian optimization (BO) is as follows. 
Since evaluating the objective function $f$ for configuration $\bm{x}$ is very expensive, it approximates $f$ using a probabilistic surrogate model $M:p(f|D)$ that is much cheaper to evaluate.
% The surrogate model can output posterior predictions for each configuration based on the previous evaluation results.
% Given a configuration $\bm{x}$, the surrogate model $M$ outputs the posterior predictive distribution at $\bm{x}$, that is,
% $f(\bm{x}) \sim \mathcal{N}(\mu_{M}(\bm{x}), \sigma^2_{M}(\bm{x}))$.
In the $n^{th}$ iteration, BO methods iterate the following three steps: (1) use the surrogate model $M$ to select a configuration that maximizes the acquisition function $\bm{x}_{n}=\arg\max_{\bm{x} \in \mathcal{X}}a(\bm{x}; M)$,
% \footnote{\color{red} Ji's comment: it makes more sense to me if we use $M_n$ 
% here other than $M$ since $M$ needs to update iterative; otherwise the subscript $n$ in $x_n$ does not make sense here. So is $D$ in the following.} 
where the acquisition function is used to balance the exploration and exploitation; (2) evaluate the configuration $\bm{x}_{n}$ to get its performance $y_{n}$; (3) add this measurement $(\bm{x}_{n}, y_{n})$ to the observed measurements $D = \{(\bm{x}_1, y_1),...,(\bm{x}_{n-1}, y_{n-1})\}$, and refit the surrogate $M$ on the augmented $D$.
Popular acquisition functions include EI~\cite{jones1998efficient}, PI~\cite{snoek2012practical}, UCB~\cite{srinivas2010gaussian}, etc.
Due to the ever-increasing evaluation cost, several researches~\cite{wang2013bayesian,falkner2018bohb} reveal that vanilla BO methods with complete evaluations fail to converge to the optimal configuration quickly.

% Expected improvement (EI)~\cite{jones1998efficient} is a common acquisition function:
% \begin{equation}
% \begin{small}
% \label{eq_ei}
% a(\bm{x}; M)=\int_{-\infty}^{\infty} \max(y^{\ast}-y, 0)p_{M}(y|\bm{x})dy,
% \end{small}
% \end{equation}
% where $y^{\ast}$ is the best observed performance in $D$, i.e., $y^{\ast}=\min\{y_1, ..., y_n\}$, and $M$ is the probabilistic surrogate model. 
% By maximizing this EI function $a(\bm{x}; M)$ over the hyperparameter space $\mathcal{X}$, BO methods can find a configuration with the largest EI value to evaluate in each iteration.

\vspace{-0.5em}
\subsection{Hyperband}
\label{sec:hb}
To address the issue in vanilla BO methods, Hyperband (HB)~\cite{li2018hyperband} proposes to speed up configuration evaluations by early stopping the badly-performing configurations. 
It has the following two loops:

(1) {\em Inner loop: successive halving. }
HB extends the original successive halving algorithm (SHA)~\cite{jamieson2016non}, which serves as a subroutine in HB, and here we also refer to it as SHA. 
SHA is designed to identify and terminate poor-performing hyper-parameter configurations early, instead of evaluating each configuration with complete training resources, thus accelerating configuration evaluation.
Given a kind of training resource (e.g., the number of iterations, the size of training subset, etc.), 
SHA first evaluates $n_1$ hyper-parameter configurations with the initial $r_1$ units of resources each, and ranks them by the evaluation performance.
Then it promotes the top $1 / \eta$ configurations to continue its training with $\eta$ times larger resources (usually $\eta=3$), that's, $n_2=n_1*\eta^{-1}$ and $r_2=r_1*\eta$, and stops the evaluations of the other configurations in advance. 
This process repeats until the maximum training resource $R$ is reached.
% The ASHA procedure is a kind of resource allocation, which allocates different training resources to each configuration based on their performance.
Figure~\ref{fig:hb_sh} gives a concrete example of SHA.

(2) {\em Outer loop: the choice of $r_1$ and $n_1$.}
Given some finite budget $B$ for each bracket, the values of $r_1$ and $n_1=\frac{B}{r_1}$ should be carefully chosen because a small initial training resource $r_1$ with a large $n_1$ may lead to the elimination of good configurations in SHA iterations by mistake.
There is no prior whether we should use a smaller initial training resource $r_1$ with a larger $n_1$, or a larger $r_1$ with a smaller $n_1$. 
HB addresses this problem by enumerating several feasible values of $r_1$ in the outer loop, where the inner loop corresponds to the execution of SHA. 
Table~\ref{tb:hb_evaluations} shows a concrete example about the number of evaluations and their corresponding training resources in an iteration of HB, where each column corresponds to the results of inner loop (i.e., one iteration of SHA with different $r_1$s).
For example, the first column ``Bracket-1'' of Table~\ref{tb:hb_evaluations} corresponds to the execution process of SHA in Figure~\ref{fig:hb_sh}.
Note that, {\em the HB iteration will be called multiple times until the tuning budget exhausts}.

\iffalse
\begin{algorithm}[tb]
  % \algsetup{linenosize=\normalsize}
  \scriptsize
  \caption{Pseudo code for Hyperband.}
  \label{algo:hb}
  \textbf{Input}: maximum number of resources that can be allocated to a single hyperparameter configuration $R$, the discard proportion $\eta$, and hyperparameter space $\mathcal{X}$.
  
  \begin{algorithmic}[1]
  \STATE Initialize $s_{max}=\lfloor log_{\eta}(R) \rfloor$, $B = (s_{max}+1)R$.
  \FOR{$s \in \{s_{max}, s_{max}-1, ..., 0\}$}
      \STATE $n_1 = \lceil \frac{B}{R}\frac{\eta^s}{s+1} \rceil$, $r_1 = R\eta^{-s}$.
      \STATE sample $n_1$ configurations from $\mathcal{X}$ randomly.
      \STATE execute the SHA with the $n_1$ configurations and $r_1$ as input (the inner loop).
  \ENDFOR
  \STATE \textbf{return} the configuration with the best evaluation result.
\end{algorithmic}
\end{algorithm}
\fi

\begin{table}[tb]
  \centering
  \small
  \caption{The values of $n_i$ and $r_i$ in the HB evaluations, where $R = 27$ and $\eta = 3$. Each column displays an inner loop (SHA process). The pair ($n_i$, $r_i$) in each cell means there are $n_i$ configuration evaluations with $r_i$ units of training resources.
  Taking the first column ``Bracket-1'' as an example, the evaluation process corresponds to the iteration of SHA in Figure 2.
  It starts with $n_1=27$ config evaluations with $r_1=1$ unit of resources; then the top $n_2=9$ evaluations continue with $r_2=3$; then $n_3=3$ and $r_3=9$; finally, only $n_4=1$ config gets the maximum resource $r_4=27$.
  }
  \vspace{-1em}
  \resizebox{0.75\columnwidth}{!}{
  \begin{tabular}{|c|lll|lll|lll|lll|} 
    \hline
    \multirow{1}{*}{} &
    \multicolumn{3}{l|}{Bracket-1} &
    \multicolumn{3}{l|}{Bracket-2} &
    \multicolumn{3}{l|}{Bracket-3} &
    \multicolumn{3}{l|}{Bracket-4} \cr
    $i$ & $n_i$ & & $r_i$ &
            $n_i$ & & $r_i$ &
            $n_i$ & & $r_i$ &
            $n_i$ & & $r_i$ \\

    \hline
    $1$ & \underline{$27$} & & $1$ & 
            \underline{$12$} & & $3$ & 
            \underline{$6$} & & $9$ & 
            $\bm{4}$ & & $27$\\
    $2$ & \underline{$9$} & & $3$ & 
            \underline{$4$} & & $9$ & 
            $\bm{2}$ & & $27$ & 
             & & \\
    $3$ & \underline{$3$} & & $9$ & 
            $\bm{1}$ & & $27$ & 
            & & & & &\\
    $4$ & $\bm{1}$ & & $27$ & 
            & & & & & & & & \\
    \hline
  \end{tabular}
  }
  \label{tb:hb_evaluations}
  \vspace{1em}
\end{table}

{\bf Definitions. }
We refer to SHA with different initial training resources -- $r_1$s as {\em brackets} (See Table~\ref{tb:hb_evaluations}), and the evaluations with certain units of training resources as {\em resource level}. 

{\bf Partial Evaluation Design Issue in HB. }
Since HB-style methods~\cite{li2018hyperband,falkner2018bohb,li2021mfes} own excellent features, such as flexibility, scalability, and ease of parallelization, we build our framework based on HB.
HB consists of multiple brackets (i.e., SHA procedures), and each of them requires an $r_1$ as input.
% Without any priors, it is difficult to choose a proper $r_1$ for each bracket. 
HB enumerates several feasible values of $r_1$, and executes each corresponding bracket sequentially and repeatedly.
Bracket-$i$ is equipped with $r_1 = \eta ^ {i-1}$ units of initial training resources, so each bracket corresponds to a kind of partial evaluation design. 
When digging deeper into the HB framework, we observe the ``precision vs. cost'' tradeoff caused by the selection of bracket (i.e., each kind of partial evaluation design) as follows:
(1) The partial evaluation with a small $r_1$ implies that few training resources are allocated, and this may incur a larger number of inaccurate promotions in SHA, i.e., poor configurations are promoted to the next resource level, and good configurations are terminated by mistake due to the low fidelity.
(2) As $r_1$ becomes large, the partial evaluation design has the risk of high evaluation cost but diminishing returns from precision improvements. 
While HB tries each bracket sequentially and repeatedly, it is inevitable that it wastes evaluation cost when applying a large number of inappropriate brackets during optimization.
% Based on this observation, trying each bracket (i.e., each kind of partial evaluation design) sequentially and repeatedly as HB does is quite inefficient.
To develop an efficient tuning system, we need to revisit the HB pipeline and answer the following question:
{\em 
Can we automatically learn the right level of resource allocation (i.e., proper partial evaluation design) that balances the ``precision vs. cost'' tradeoff well?
}
In Section~\ref{sec:bracket_selection} we describe our bracket selection based solution to this problem.

\begin{figure}[tb]
	\centering
		\scalebox{0.8}[0.8] {
		\includegraphics[width=1\linewidth]{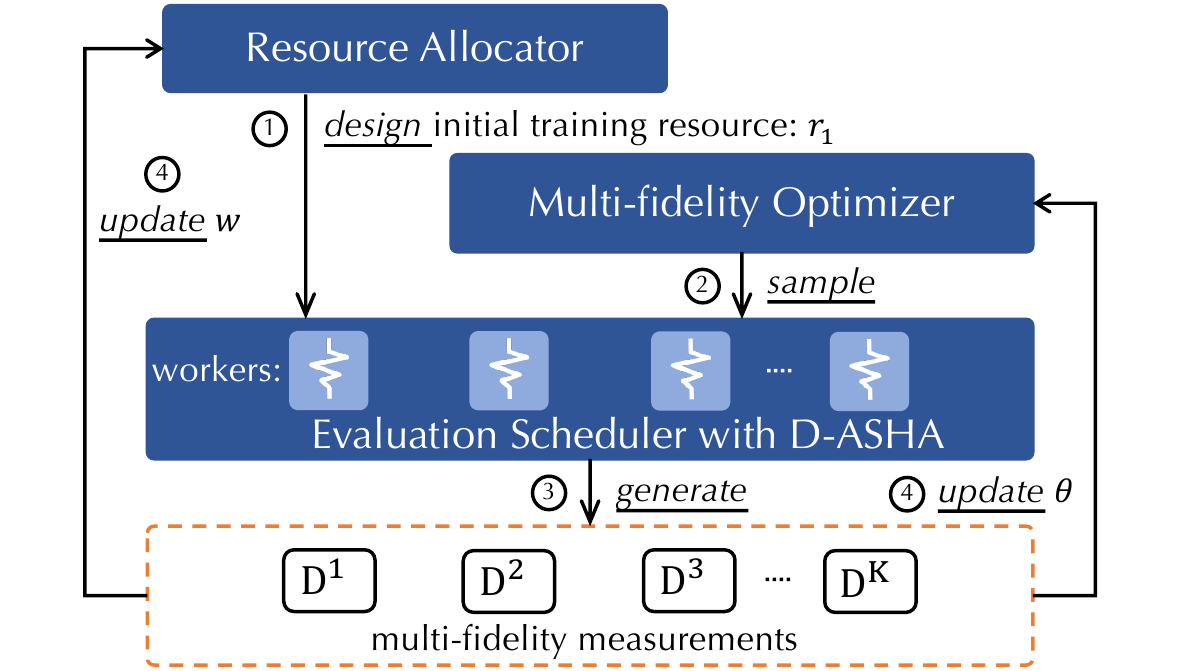}
         }
	\vspace{-0.5em}
	\caption{The framework of \sys.
	}
    \label{fig:framework}
\end{figure}

\section{Proposed Framework}
\label{sec:proposed_framework}
In this section, we first give the overview of the proposed framework, and then describe three core components that are designed to accommodate the aforementioned three issues in Section~\ref{sec:intro}.

{\em \underline{Framework Overview.}} 
The proposed framework -- \sys takes the tuning task and time budget as inputs, and outputs the best configuration found in the search process.
\sys has three components: resource allocator (Section~\ref{sec:bracket_selection}), evaluation scheduler (Section~\ref{sec:dasha}), and multi-fidelity optimizer (Section~\ref{sec:mfs}). 
It is an iterative framework that will repeat until the given budget exhausts.
Figure~\ref{fig:framework} illustrates an iteration of \sys, with 
four concrete steps.
The resource allocator selects the initial training resources $r_{1}$ when evaluating configurations (Step 1), which directly determines partial evaluation design.
% Based on a policy, the resource allocator outputs the initial training resources $r_{1}$ when evaluating configurations (Step 1), which directly determines partial evaluation design.
Then the multi-fidelity optimizer will sample a configuration from the search space for each idle worker (Step 2).
The evaluation scheduler then evaluates these configurations with the corresponding training resources in parallel (Step 3).
Finally, based on the multi-fidelity results from parallel evaluations, \sys updates the parameters in resource allocator and multi-fidelity optimizer (Step 4), respectively.

{\em \underline{Basic Setting: Measurements and Base Surrogates.}}
Due to the flexibility and scalability of HyperBand (HB)~\cite{li2018hyperband,li2018massively,falkner2018bohb,abohb}, we build our framework on HB. 
Then we collect the results from evaluations with different resource levels, and we refer to them as ``measurements''. 
According to the number of training resources used by the evaluations, we can categorize the measurements into $K$ groups: $D_1, ..., D_K$, where $K=\lfloor \log_{\eta}(R) \rfloor+1$, $\eta$ is the discard proportion in HB, $R$ is the maximum training resources for evaluation, and typically $K$ is less than $7$.
The measurement ($\bm{x}$, $y$) in each group $D_{i}$ with $i\in [1:K]$ is obtained by evaluating configuration $\bm{x}$ with $r_i=\eta^{i-1}$ units of training resources.
Thus $D_K$ denotes the high-fidelity measurements from the complete evaluations with the maximum training resources $r_K=R$, and $D_{1:K-1}$ denote the low-fidelity measurements from the partial evaluations.
In \sys, we build $K$ base surrogates: $M_{1:K}$, where surrogate $M_{i}$ is trained on the group of measurements $D_i$. 
In the following sections, we introduce the design of each component.

\vspace{-0.5em}
\subsection{Resource Allocation with Bracket Selection}

\textbf{\em \underline{Challenge. }}
\label{sec:bracket_selection}
The resource allocator aims to design the proper partial evaluations automatically.
As stated in Section~\ref{sec:hb}, the optimal bracket (i.e., the optimal initial training resources that balance the ``precision vs. cost'' trade-off well) minimizes the evaluation cost while keeping a high precision of partial evaluations. 
{\em We need to automatically deal with this trade-off.}
The resource allocator needs to identify the optimal bracket in HB, where each bracket corresponds to a type of initial resource design for partial evaluation.
% Therefore, we need to find the minimal training resource to execute bracket so that good configurations can be distinguished accurately.

\noindent
\textbf{\em \underline{Solution Overview. }}
We adopt the ``trial-and-error'' paradigm to identify the optimal bracket in an iterative manner.
In each iteration, it iterates the following three steps: (1) we first select a bracket (i.e., partial evaluation design involving $n_1$ configurations with $r_1$ initial training resources) based on parameters $\bm{w}$; (2) once the $i^{th}$ bracket is chosen, we execute this bracket; (3) based on the measurements from these evaluations, we could update the parameters $\bm{w}$.
{\em For Step 1, in the beginning, we select brackets by round-robin with three times (as initialization)};
then we sample a bracket using parameters $\bm{w}$, where each $w_i$ with $i\in[1:K]$ indicates the probability of this bracket being the optimal one.
For Step 3, we propose a two-stage technique to calculate $\bm{w}$ that balances the above trade-off. 
In the first stage, we learn a parameter $\theta_i$ for each bracket, where $\theta_i$ is proportional to the precision of evaluations with the training resources $r_i$.
In the second stage, we multiply each $\theta_i$ with a coefficient $c_i$ to obtain the final $w_i$. 
This coefficient is inversely proportional to the training resources in the partial evaluation; in this way, the strategy tends to choose the bracket with small training resources.
{\em By the multiplication between $c_i$ and $\theta_i$ ($w_i=c_i \cdot \theta_i$), we could balance the ``precision vs. cost'' trade-off in partial evaluations.}

To measure precision, we focus on the partial orderings of measurements among different resource levels.
If configuration $\bm{x}_1$ performs better than $\bm{x}_2$ when the training resource is $r$, given the complete training resource $\bm{x}_1$ still outperforms $\bm{x}_2$, indicating that the partial evaluations with $r$ units of training resources are accurate, so we can utilize this to measure the precision of evaluations.
To implement this, we utilize the predictions of base surrogate $M_i$ built on $D_i$, and compare the predictive rankings of configurations with the rankings in $D_K$.
% The real value of the prediction does not matter and we care about the partial orderings (i.e., rankings) over configurations. 
For base surrogates $M_{1:K-1}$, we define the ranking loss as the number of miss-ranked pairs as follows:
\begin{equation}
    \scriptsize
    \mathbb{L}(M_i) = \sum_{j=1}^{N_K}\sum_{k=1}^{N_K}\mathds{1}((M_i(\bm{x}_j) < M_i(\bm{x}_k) \otimes (y_j < y_k)),
    \label{eq:ordering}
\end{equation}
where $\otimes$ is the exclusive-or operator,
% \footnote{\color{red} Ji's comment: we need an example here to illustrate $\otimes$. I do not know if this notation follows a conventional definition. In machine learning people usually use $\lor$ to denote the ``or'' logical operator}
$N_K = |D_K|$, and $(\bm{x_i}, y_i)$ is the measurement in $D_K$.
For the base surrogate $M_K$ trained on $D_K$ directly, we adopt 5-fold cross-validation to calculate its $\mathbb{L}(M_K)$.
Further we define each $\theta_i$ as the probability that base surrogate $M_i$ has the least ranking loss.
Concretely, we use Markov chain Monte Carlo (MCMC) to learn $\bm{\theta}$ by drawing $S$ samples: $l_{i,s} \sim \mathbb{L}(M_i)$ for $s=1,...,S$ and each surrogate $i=1,...,K$, and calculating
\begin{equation}
    \scriptsize
    \theta_i=\frac{1}{S}\sum_{s=1}^{S}\mathds{1}\left(i=\mathop{\arg\min}_{i'}l_{i',s}\right).
\end{equation}
To obtain $\bm{c}$, in \sys we simply apply the inverse of the corresponding training resources, i.e., $c_i = 1 / r_i$. Finally, we normalize the raw $\bm{w} = \bm{c} \circ \bm{\theta}$ to obtain the final $\bm{w}$, where $\sum_i{w_i}=1$.

\vspace{-1.em}
\subsection{Asynchronous Evaluation Scheduling}
\label{sec:dasha}
\begin{figure}[tb]
	\centering
		\scalebox{1.}[1.] {
		\includegraphics[width=1\linewidth]{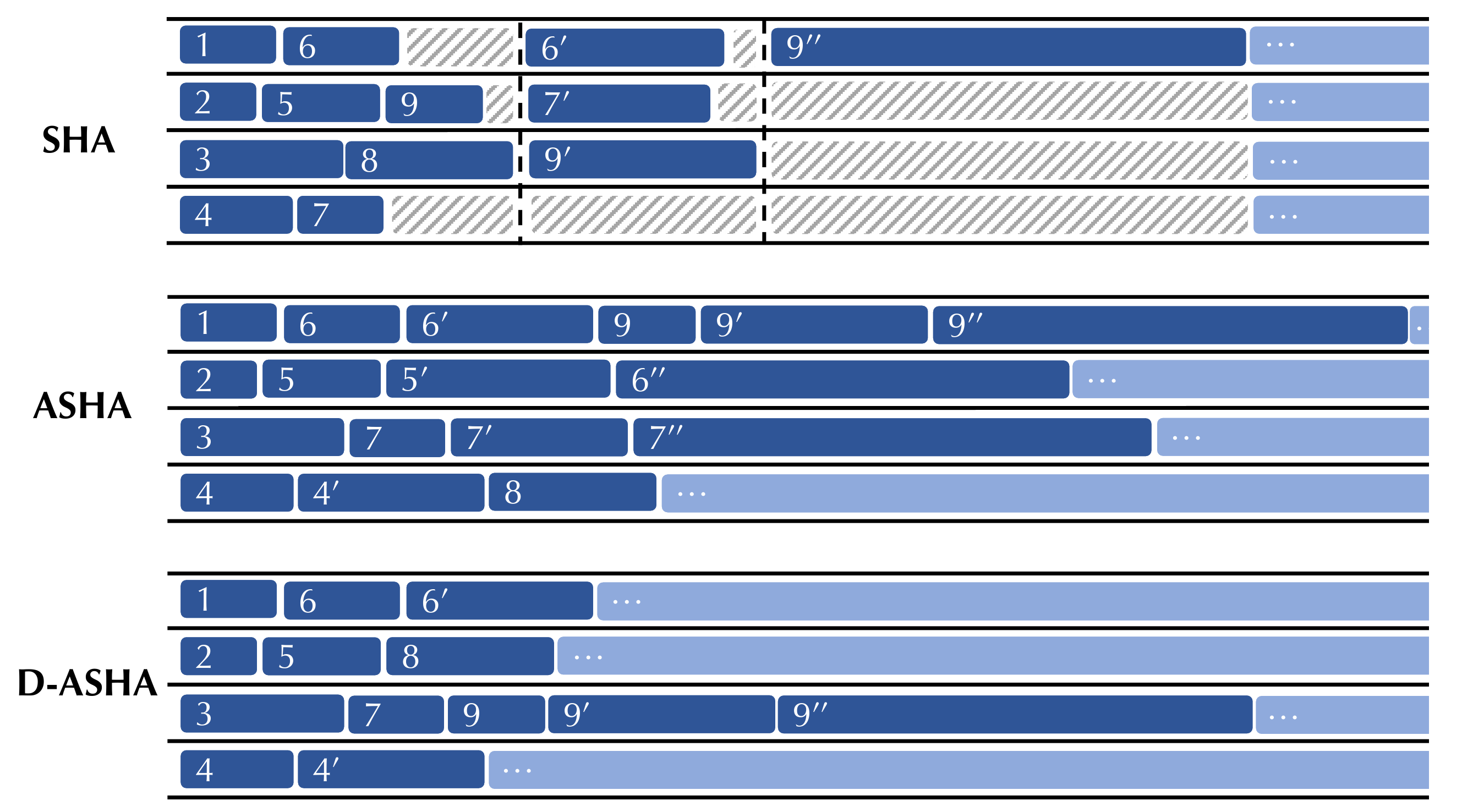}
         }
    \vspace{-2.em}
	\caption{Three scheduling mechanisms on a real-world case, where each row corresponds to a worker, and the ranking of configurations are $\bm{x}_3$, $\bm{x}_8$, $\bm{x}_2$, $\bm{x}_1$, $\bm{x}_4$, $\bm{x}_5$, $\bm{x}_6$, $\bm{x}_7$, $\bm{x}_9$ (latter is the better). $i$' refers to promoted evaluation of configuration $\bm{x}_i$.
	Each deep-blue block with $i$ corresponds to the evaluation process of $\bm{x}_i$; the light-blue blocks represent the evaluations for other iterations of SHA procedures; the striped areas in SHA refer to no evaluations for workers.}
    \label{fig:parallel_strategy}
\end{figure}

\textbf{\em \underline{Challenge. }}
In this section, we introduce the distributed scheduling mechanism in the evaluation scheduler. 
SHA~\cite{jamieson2016non} promotes the top $1 / \eta$ configurations to the next resource level until all configurations in the current level have been evaluated (synchronization barrier). 
Due to the synchronous design, which often leads to the straggler issue, the ineffective use of computing resources in SHA is inevitable.
ASHA~\cite{li2020system} is able to remove these issues associated with synchronous promotions by incurring a number of inaccurate promotions (See Figure~\ref{fig:parallel_strategy}), i.e., configurations that fall into the top $1 / \eta$ early but are not in the actual top $1 / \eta$ of all configurations. 
However, this frequent and inaccurate promotion could hamper the sample efficiency when utilizing the parallel and distributed resources, i.e., ASHA may spend lots of training resources on evaluating the less promising configurations.
\textit{Therefore, we need an efficient scheduling method which pursues high sample efficiency while keeping the advantage of asynchronous mechanism}.
% Evidently, there still remains space to develop an efficient scheduling mechanism.

% Pseudo Code for D-ASHA.
\begin{algorithm}[tb]
  % \algsetup{linenosize=\normalsize}
  \small
  \caption{Pseudo Code for D-ASHA.}
  \label{algo:delayed_asha}
  \KwIn{initial training resource $r_{1}$, maximum resource $R$, discard proportion $\eta$.}
  
  \SetAlgoLined
  \textbf{Function} $\operatorname{D-ASHA()}$:
  
  \quad \  $\bm{x}, r_x = \operatorname{get\_job()}$;
  
  \quad \  Assign a job with configuration $\bm{x}$ and resource $r_x$ to a free worker.
  
  \textbf{Function} $\operatorname{get\_job()}$:
  
  \quad \  \textit{// Check if we need to promote configurations.}
  
  \quad \ \textbf{for} $k = \lfloor log_{\eta}(R) \rfloor, ..., 2, 1$, \textbf{do}
  
    \quad \ \quad \  \textit{// $D_k$ refers to measurements of resource level $k$}.
    
    \quad \ \quad \  Configuration candidates $\mathit{C} = \{{\bm x}\  \text{for} \ {\bm x} \in$ top $1 / \eta$ configurations in $D_k$ if $\bm{x}$ has not been promoted\}
    
    \quad \ \quad \ \textbf{if} $|D_k| / (|D_{k+1}| + 1) \ge \eta$ and $|\mathit{C}| > 0$, \textbf{then}
    
    \quad \ \quad \ \quad \ \textbf{return} $\mathit{C}[0], \eta^{k}$ 
    
    \quad \ \quad \ \textbf{end if}
    
    \quad \ \textbf{end for}

  \quad \ Sample a configuration $\bm{x}$ based on the multi-fidelity optimizer.
  
  \quad \ \textbf{return} $\bm{x}, r_{1}$
\end{algorithm}

\noindent
\textbf{\em \underline{Delayed ASHA}. }
To alleviate this issue, we propose a variant of ASHA --- delayed ASHA (abbr. D-ASHA), which uses a delay strategy to decrease inaccurate promotions and still preserves the asynchronous scheduling mechanism.
% To address the promotion issue in ASHA, we propose a variant of ASHA --- delayed ASHA (abbr. D-ASHA) as shown in Algorithm ~\ref{algo:delayed_asha}.
% While achieving faster convergence compared with ASHA, D-ASHA can obtain higher sample efficiency when utilizing the parallel resources (See D-ASHA in Figure~\ref{fig:parallel_strategy}).
Instead of promoting each configuration that is in the top $1 / \eta$ of all previously-evaluated configurations, D-ASHA promotes configurations to the next level if (1) the configuration is in the top $1 / \eta$ of configurations, and (2) the number of collected measurements $|D_k|$ with current resource level should be $\eta$ times larger than the number of the next level's $|D_{k+1}|$ if promoted (Lines 9-10 in Algorithm~\ref{algo:delayed_asha}). 
The inaccurate promotions (in Cond.1) arise from a small number of observed measurements in $D_k$ with current resource level.
The condition 2 ensures that $|D_k|$ should be larger than a threshold $\eta|D_{k+1}|$, i.e., $|D_k| / (|D_{k+1}| + 1) \ge \eta$.
In this way, the delay strategy could prevent the frequent promotion issue in ASHA, and further improve the sample efficiency.
Figure 4 gives a concrete real-world example to explain this design.
Algorithm~\ref{algo:delayed_asha} provides the formulated description about D-ASHA.
Additionally, if no promotions are possible, D-ASHA requests a new configuration from the multi-fidelity optimizer (provided in Algorithm~\ref{algo:paralllel_sample}) and adds it to the base level (Lines 13-14), so that more configurations can be promoted to the upper levels.

\vspace{-1em}
\subsection{Multi-fidelity Configuration Sampling}
\label{sec:mfs}
\textbf{\em \underline{Challenge. }}
There are various advancements in the design of Bayesian optimization (BO) methods. While those algorithms differ in the execution process, \textit{a flexible tuning system should contain an optimizer module that allows us to plug in different hyper-parameter tuning optimizers easily. }
In addition, since most BO based methods are intrinsically sequential, it is impractical to modify each possible algorithm to support parallel scenarios case by case.
Thus, we need an algorithm-agnostic framework to extend different sequential optimizers to support parallel evaluations in both  sync/asynchronous settings.

\noindent
\textbf{\em \underline{Optimizer Design. }}
To tackle the first challenge, we provide a generic optimizer abstraction for configuration sampling in \sys.
It includes 1) the data structure to store multi-fidelity measurements: $D_1$, ..., $D_K$, and 2) the \texttt{fit} and \texttt{predict} APIs for surrogate model.
This abstraction enables convenient support/implementation of different configuration sampling algorithms (e.g., random search, Bayesian optimization, multi-fidelity optimization, etc.).
For the second challenge, we further propose an algorithm-agnostic sampling framework to support asynchronous and synchronous parallel evaluations conveniently without any ad-hoc modifications.

\noindent
{\em \underline{(Multi-fidelity Optimizer.)}}
Multi-fidelity methods~\cite{kleinfbhh17,kandasamy2017multi,poloczek2017multi,hu2019multi,sen2018noisy,wu2019practical,li2020efficient} have achieved success in hyper-parameter tuning.
Meanwhile, it produces multi-fidelity measurements which can help determine the optimal bracket for evaluation.
% In practice, we observe that when performing hyper-parameter tuning, the low-fidelity measurements could provide useful information about the objective function, and can be used to speed up the search process. 
In \sys, we implement a multi-fidelity optimizer by default based on MFES-HB~\cite{li2020efficient} to utilize multi-fidelity measurements, and build a multi-fidelity ensemble surrogate by combining all base surrogates:
\begin{equation}
% \begin{scriptsize}
    M_{\text{MF}} = \texttt{agg}(\{M_1,...,M_K\}; {\bm \theta});\nonumber
% \end{scriptsize}
\end{equation}
The surrogate $M_{\text{MF}}$ is used to guide the configuration search, instead of the high-fidelity surrogate $M_{K}$ only, in the framework of BO.
Concretely, we combine the base surrogates with weighted bagging, and the weights $\bm{\theta}$ are exactly the parameters obtained in Section~\ref{sec:bracket_selection}. 
Each $\theta_i$ also reflects the reliability when applying the corresponding low-fidelity information from partial evaluations with $r_i$ units of training resources to the target problem. 
Finally, the predictive mean and variance of $M_{MF}$ at configuration $\bm{x}$ are given by:
\begin{equation}
\scriptsize
\label{eq:weighted_bagging}
\begin{aligned}
    & \mu_{MF}(\bm{x})=\sum_i\theta_i\mu_{i}(\bm{x}), \quad \sigma_{MF}^2(\bm{x}) = \sum_i\theta_i^2\sigma_i^{2}(\bm{x}),
\end{aligned}
\end{equation}
where $\mu_i(\bm{x})$ and $\sigma_i^{2}(\bm{x})$ are the mean and variance predicted by the base surrogate $M_i$ at configuration $\bm{x}$.
Based on the multi-fidelity measurements, this multi-fidelity surrogate could learn the objective function well, and can be used to speed up the search process.

\begin{figure*}[htb]
	\centering
	\subfigure[CIFAR-10-valid]{
		% Requires \usepackage{graphicx}
		\scalebox{0.28}{
			\includegraphics[width=1\linewidth]{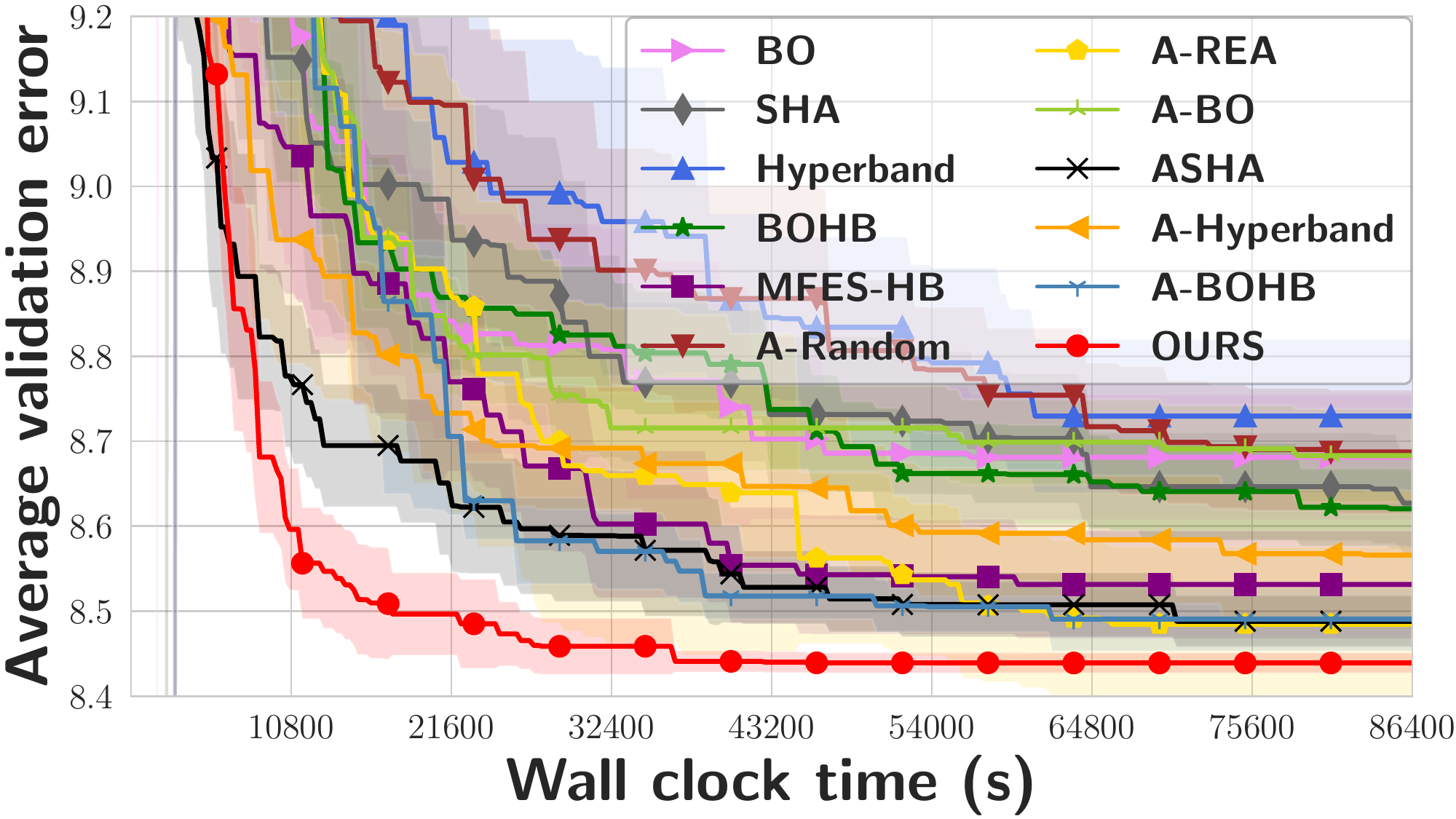}
	}}
	\subfigure[CIFAR-100]{
		% Requires \usepackage{graphicx}
		\scalebox{0.28}{
			\includegraphics[width=1\linewidth]{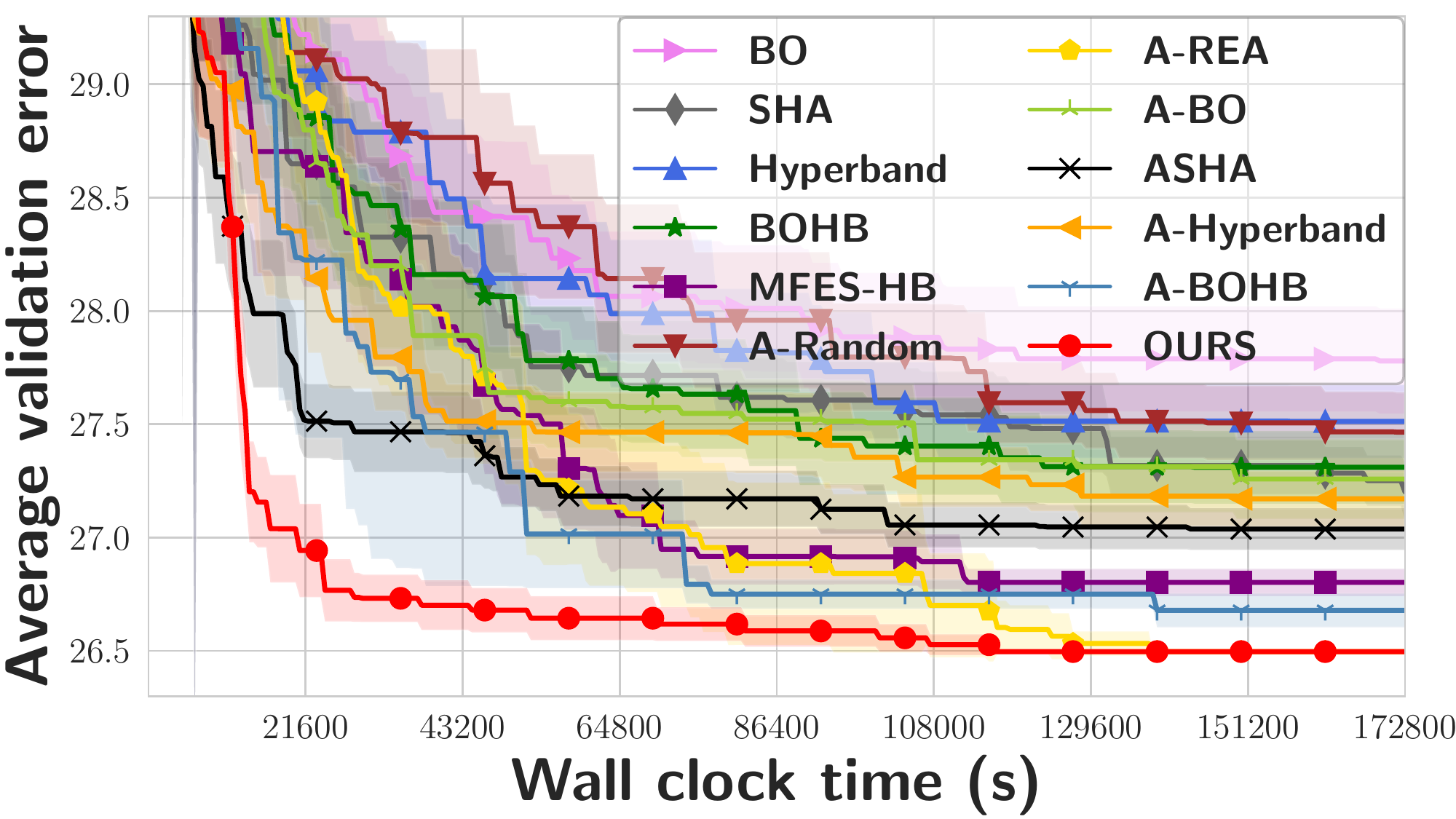}
	}}
    \subfigure[ImageNet16-120]{
		% Requires \usepackage{graphicx}
		\scalebox{0.28}{
			\includegraphics[width=1\linewidth]{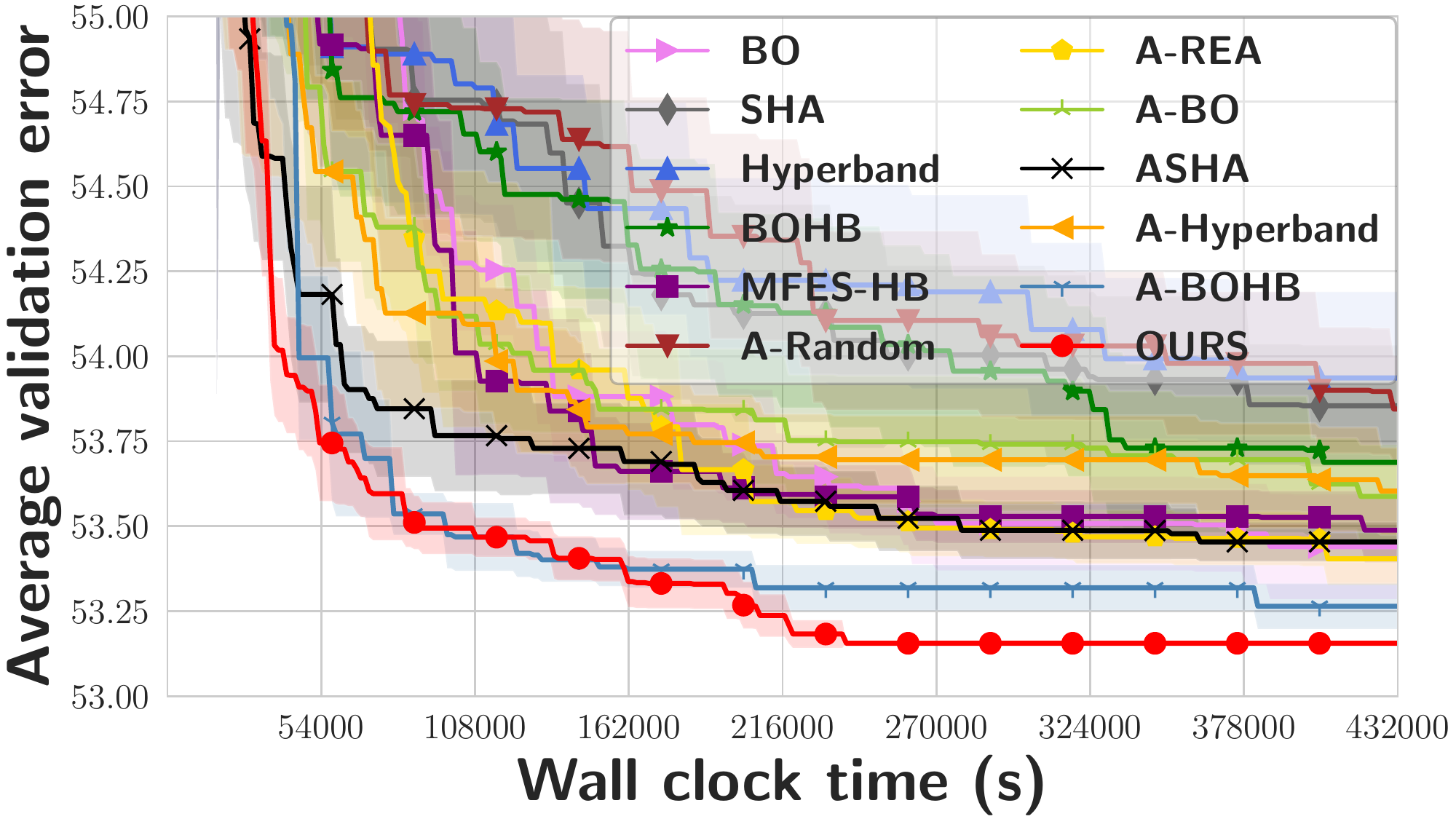}
	}}
	\vspace{-1.5em}
	\caption{Validation error (\%) of tuning architectures on three datasets based on NAS-Bench-201.}
	\vspace{-1.5em}
  \label{fig:nas}
\end{figure*}

\noindent
{\em \bf \underline{(Algorithm-agnostic Sampling.)}}
Since traditional BO methods are intrinsically sequential, we need an algorithm-agnostic framework to extend the sequential method to the sync/asynchronous parallel settings seamlessly, and meanwhile ensure convergence.
In parallel evaluations, each worker pulls a new configuration to evaluate after the previous evaluation is completed; there are pending evaluations that are not finished when sampling new configurations. 
We need to sample new configurations that are promising and far enough from the configurations being evaluated by other workers to avoid repeated or similar evaluations.
To this end, we adopt an algorithm-agnostic sampling framework, which leverages the local penalization-based strategy~\cite{gonzalez2016batch,Li2021OpenBoxAG} to deal with the above issue, where each pending evaluation is imputed with the median of performance in $D_K$.
This framework enables that each algorithm could adapt to the parallel scenarios easily.
Algorithm~\ref{algo:paralllel_sample} gives the algorithm-agnostic sampling procedure of optimizers.

\begin{algorithm}[t]
  \SetAlgoLined
  \scriptsize
  \caption{\emph{Sampling} procedure.}
  \KwIn{the hyper-parameter space $\mathcal{X}$, measurements $D_1$, $D_2$, ..., $D_K$, pending configurations $C_{\text{pending}}$ being evaluated on workers, the multi-fidelity surrogate $M_{MF}$, and acquisition function $\alpha(\cdot)$.}
  \SetAlgoLined
  \label{algo:paralllel_sample}
  calculate $\hat{y}$, the median of $\{y_i\}_{i=1}^n$ in $D_K$\;
  impute new measurements $D_{\text{new}}=\{(\bm{x}_\text{pending}, \hat{y}): \bm{x}_\text{pending} \in C_{\text{pending}}\}$\;
  refit the surrogate $M_{K}$ in $M_{MF}$ on $D_{\text{aug}}$, where $D_{\text{aug}} = D_K \cup D_{\text{new}}$, and build the acquisition function $\alpha(\bm{x}, M)$ using $M_{MF}$\;
  \textbf{return} the configuration $\bm{x}^{\ast}=\operatorname{argmax}_{\bm{x} \in \mathcal{X}}\alpha(\bm{x}, M_{MF})$.
\end{algorithm}

\section{Experiments and Results}
\label{sec:exp}
We now present empirical evaluations
of \sys. 
We first focus on the end-to-end \textit{efficiency} between 
\sys and other state-of-the-art systems.
We then study two more specific aspects, with the following two hypotheses:
\sys is more robust to the low-fidelity measurements with different scales of noises ({\em Robustness}); \sys can scale well to the scenarios with different evaluation cost and number of evaluation workers ({\em Scalability}).

\subsection{Experimental Settings}

\para{Compared Methods.} 
We compare \sys with the manual setting given by our enterprise partner and the following ten baselines.
(1) A-Random: Asynchronous Random Search~\cite{bergstra2012random} that selects random configurations to evaluate asynchronously, 
(2) BO: Batch-BO~\cite{gonzalez2016batch} that samples a batch of configurations to evaluate synchronously, 
(3) A-BO: Async Batch-BO~\cite{Li2021OpenBoxAG} that samples a batch of configurations to evaluate asynchronously,
(4) SHA: Successive Halving Algorithm~\cite{jamieson2016non} that adaptively allocates training resources to configurations with multi-stage early-stopping, 
(5) ASHA~\cite{li2020system} that improves SHA asynchronously via configuration promotion,
(6) Hyperband~\cite{li2018hyperband} that applies a bandit strategy to allocate resources dynamically based on SHA,
(7) A-Hyperband~\cite{li2020system} that extends Hyperband to asynchronous settings via ASHA,
(8) BOHB~\cite{falkner2018bohb} that combines the benefits of both Hyperband and Bayesian optimization,
(9) A-BOHB~\cite{abohb} that improves BOHB with asynchronous multi-fidelity optimizations,
(10) MFES-HB~\cite{li2021mfes} that combines Hyperband and multi-fidelity Bayesian optimization.
Note that Batch-BO, SHA, Hyperband, BOHB, and MFES-HB are synchronous methods, and the others are asynchronous ones.

\para{Tasks.}
We run experiments on the following tuning tasks:

\noindent(1) \textit{Neural Architecture Search.}
We use the NAS-Bench-201~\cite{dong2019bench} which includes offline evaluations of neural network architectures. The search space consists of 6 hyper-parameters.
The minimum and maximum number of epochs in NAS-Bench-201 are 1 and 200. HB-based methods use 4 brackets, and the default number of workers is 8. 
We evaluate \sys on three built-in datasets -- CIFAR-10-Valid, CIFAR-100, and ImageNet16-120, where the total budgets are 24, 48, and 120 hours, respectively.
We finish each method once the simulated training time reaches the given budget.

\noindent(2) \textit{Tabular Classification.}
We tune XGBoost~\cite{chen2016xgboost} on four large datasets from OpenML~\cite{vanschoren2014openml} -- Pokerhand, Covertype, Hepmass, and Higgs. 
The hyper-parameter space of XGBoost includes $9$ hyperparameters. 
For partial evaluations, we use the subset of the training set instead of using the entire set. The minimum and maximum size of subset are $1 / 27$ and $1$. 
HB-based methods use $4$ brackets, and the default number of workers is $8$. 
The time budgets for the above four datasets are $2$, $3$, $6$, and $6$ hours, respectively. Each worker is equipped with $8$ CPU cores during evaluation.

\noindent(3) \textit{Image Classification.}
We tune ResNet~\cite{he2016deep} on the image classification dataset -- CIFAR-10. 
The search space includes batch size, SGD learning rate, SGD momentum, learning rate decay, and weight decay. 
Cropping and horizontal flips are used as default augmentation operations. 
The minimum and maximum number of epochs are 1 and 200. HB-based methods use 4 brackets, and the default number of workers is 4. 
The time budget is 48 hours. Each worker has 8 CPU cores and 1 GPU during evaluation.

\begin{figure*}[htb]
	\centering
	\subfigure[Pokerhand]{
		% Requires \usepackage{graphicx}
		\scalebox{0.23}[0.23]{
			\includegraphics[width=1\linewidth]{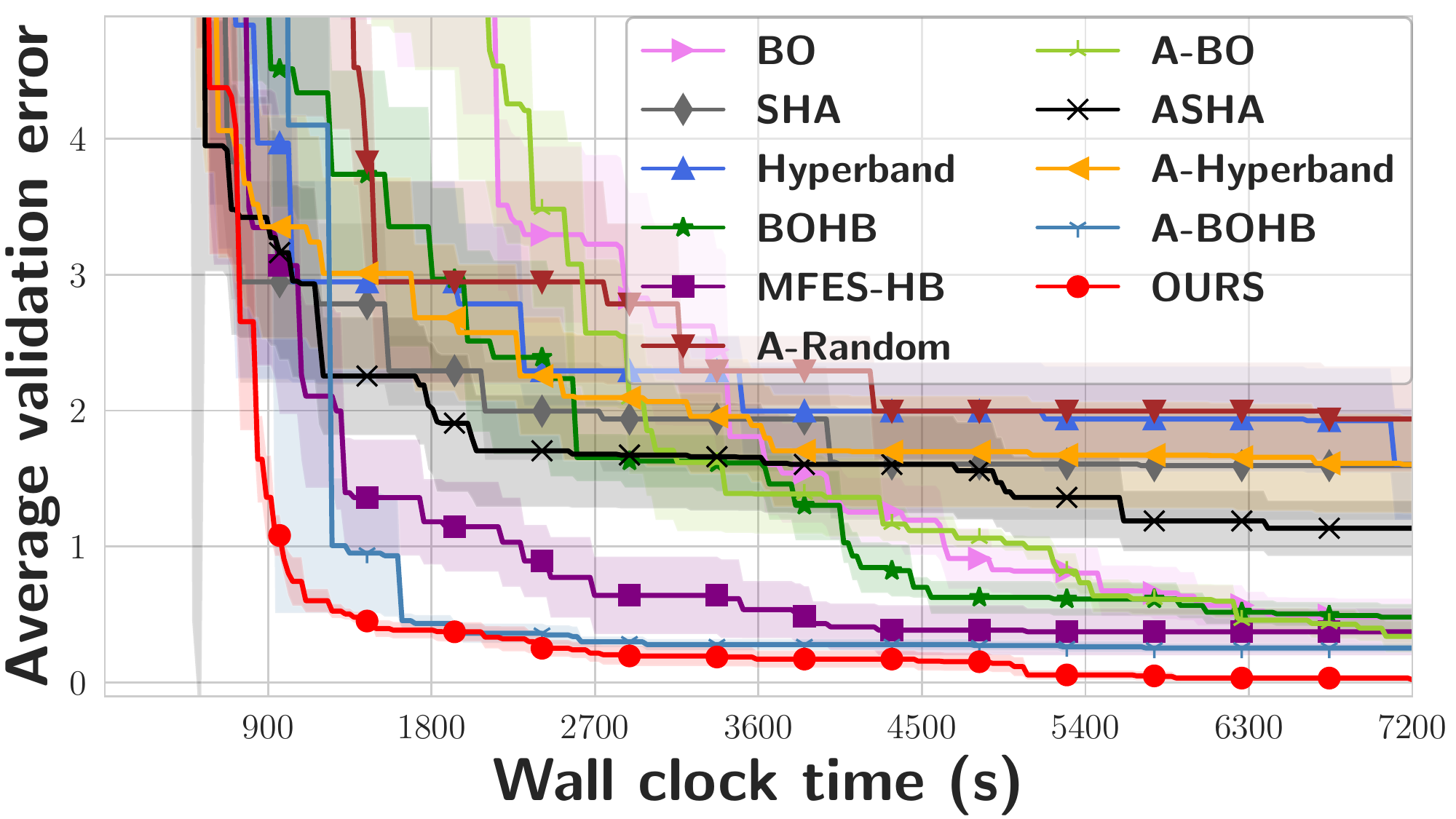}
	}}
	\subfigure[Covertype]{
		% Requires \usepackage{graphicx}
		\scalebox{0.23}[0.23]{
			\includegraphics[width=1\linewidth]{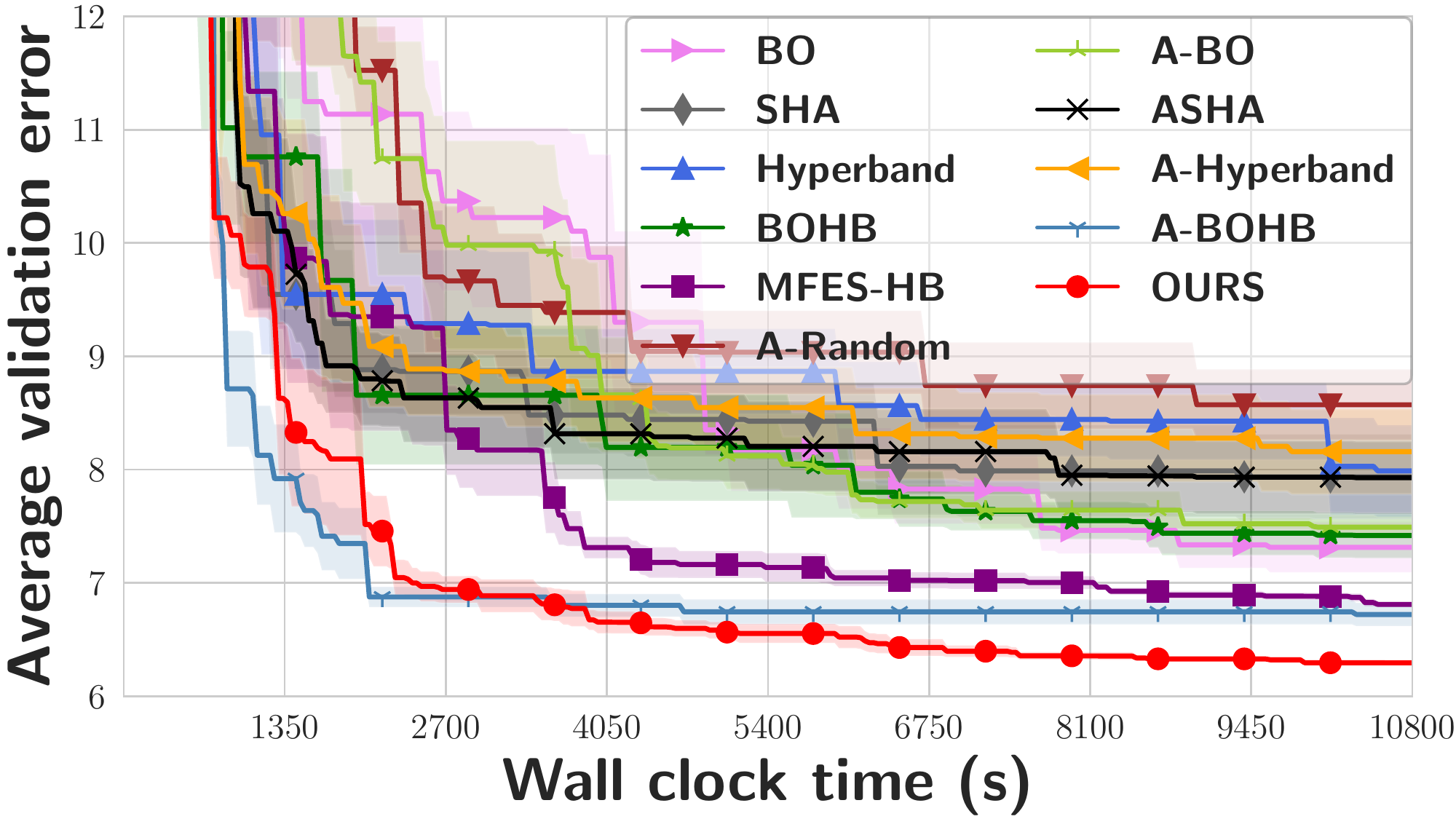}
	}}
	\subfigure[Hepmass]{
		% Requires \usepackage{graphicx}
		\scalebox{0.23}[0.23]{
			\includegraphics[width=1\linewidth]{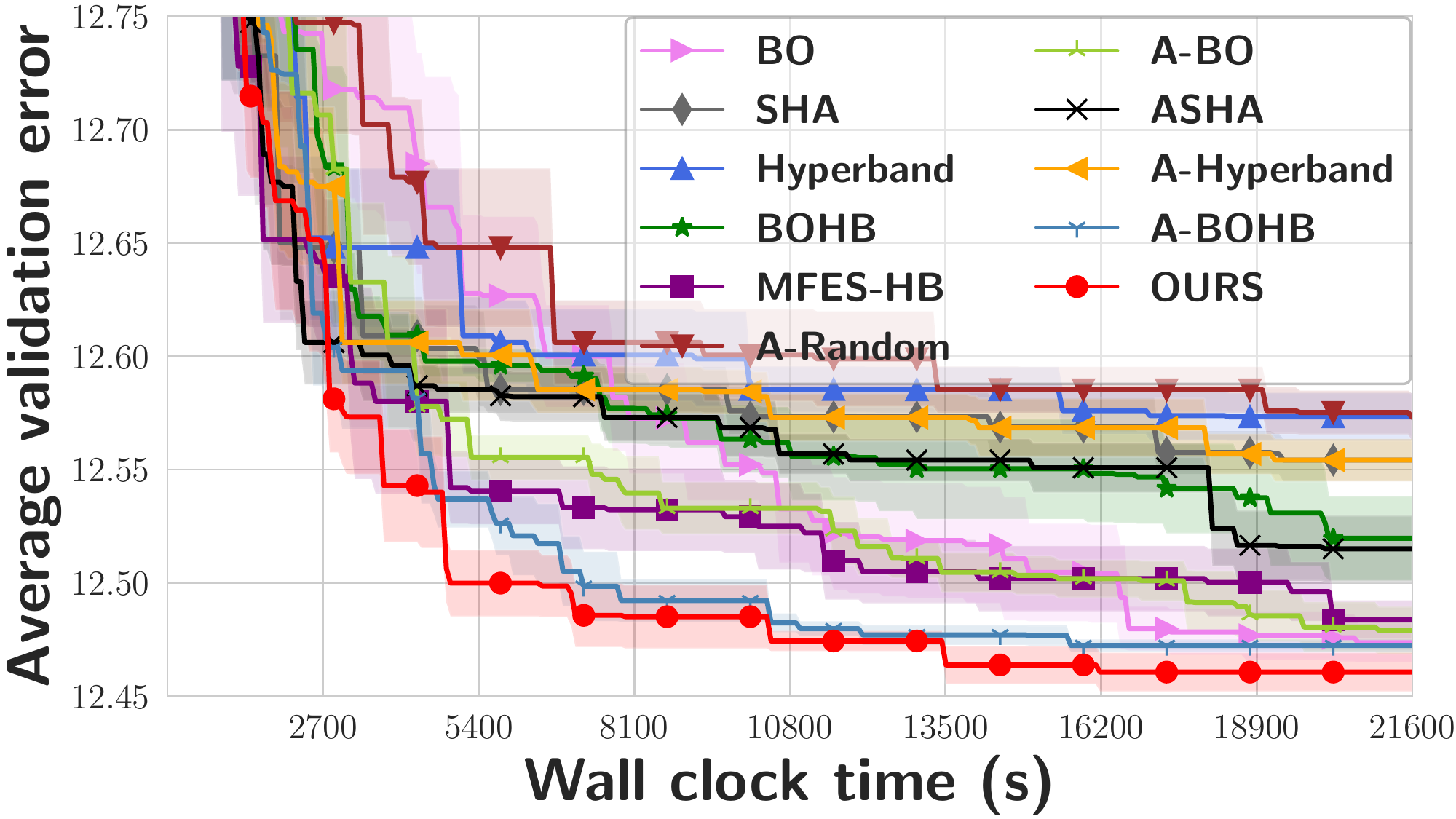}
	}}
    \subfigure[Higgs]{
		% Requires \usepackage{graphicx}
		\scalebox{0.23}[0.23]{
			\includegraphics[width=1\linewidth]{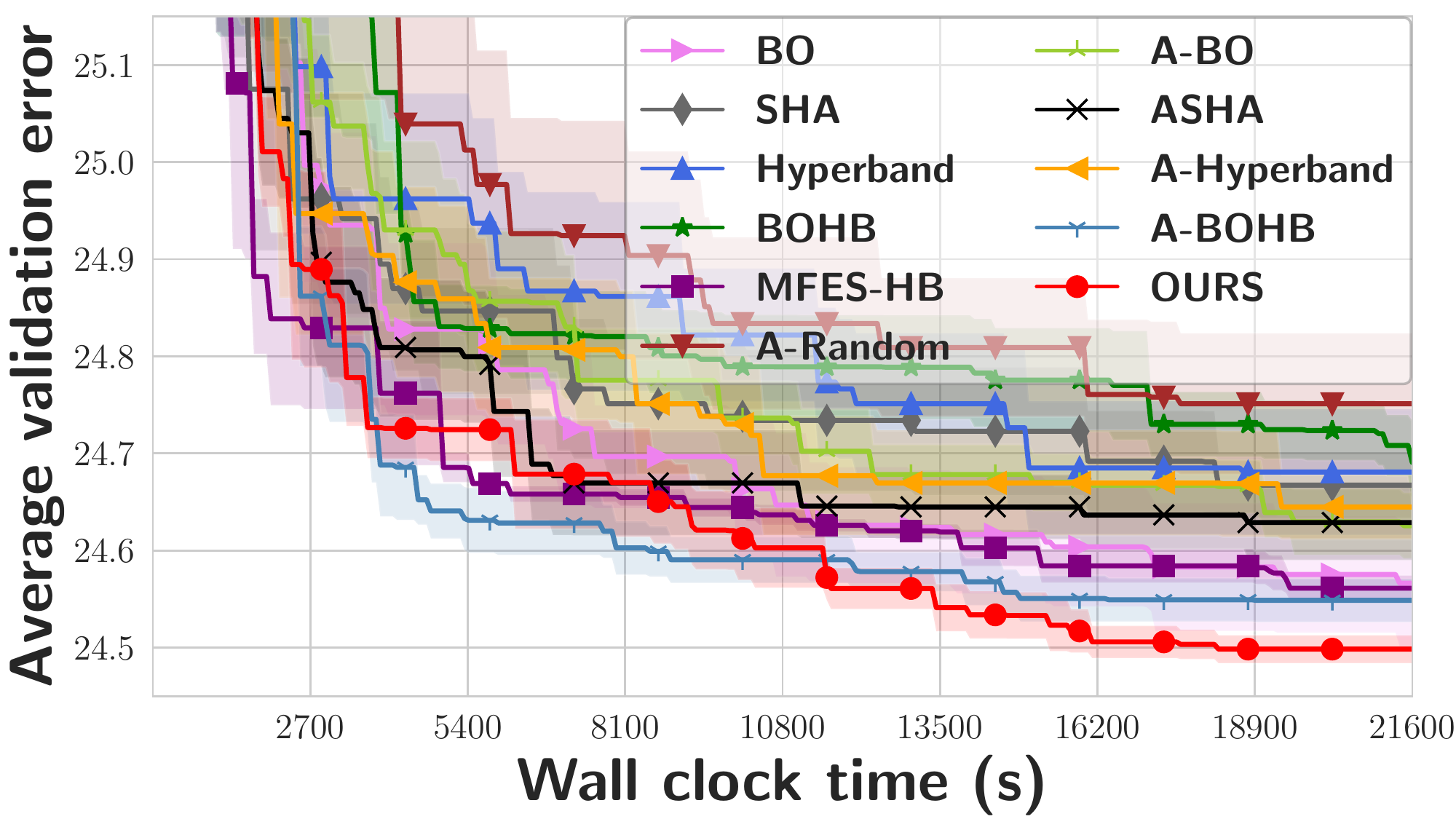}
	}}
	\vspace{-1.7em}
	\caption{Validation error (\%) of tuning XGBoost on four large datasets.}
	\vspace{-1.5em}
  \label{fig:xgboost}
\end{figure*}

\begin{figure}[tb]
	\centering
	\subfigure[LSTM on Penn Treebank]{
		% Requires \usepackage{graphicx}
		\scalebox{0.47}[0.47]{
			\includegraphics[width=1\linewidth]{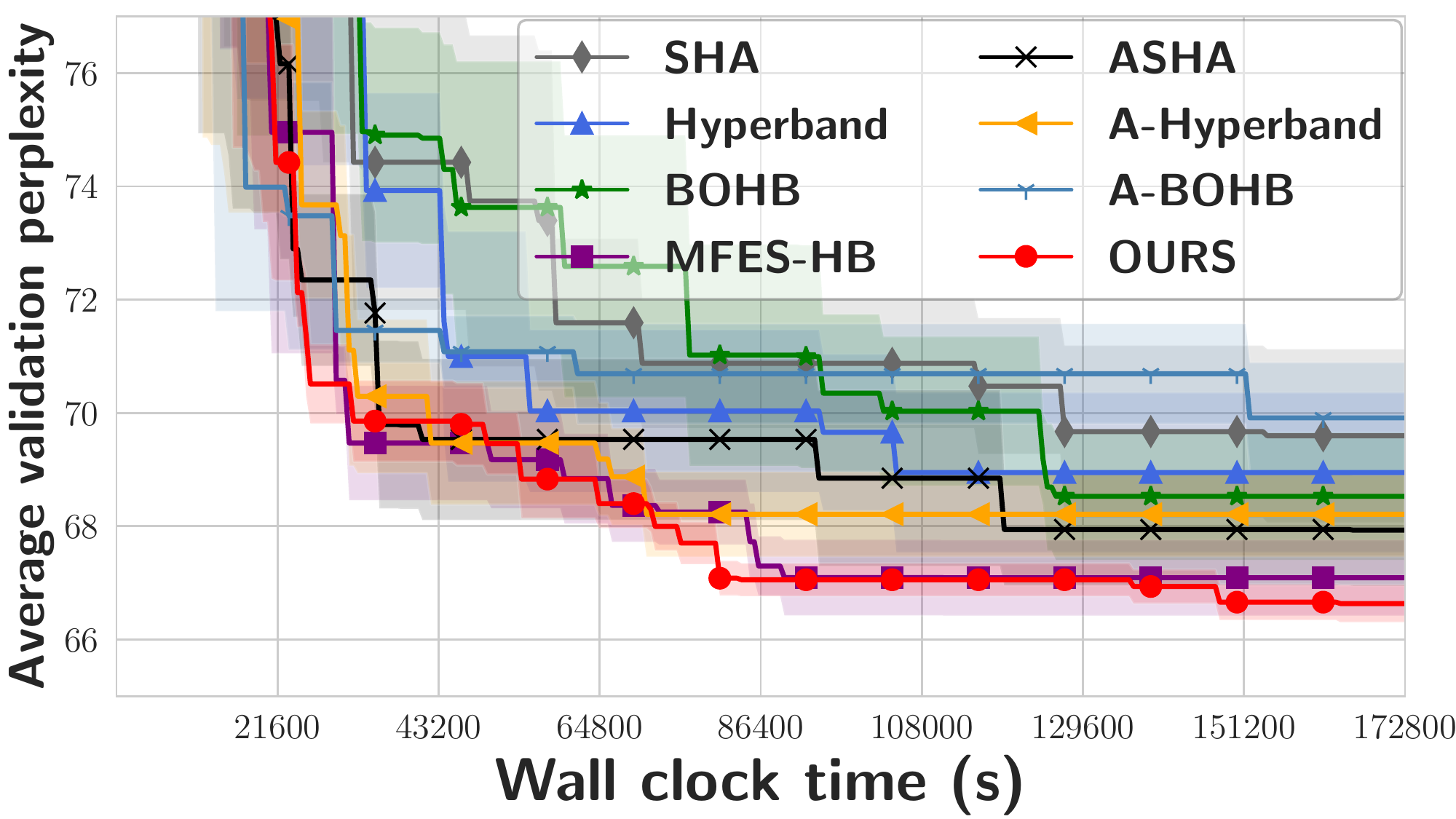}
		\label{fig:lstm}
	}}
	\subfigure[ResNet on CIFAR-10]{
		% Requires \usepackage{graphicx}
		\scalebox{0.47}[0.47]{
			\includegraphics[width=1\linewidth]{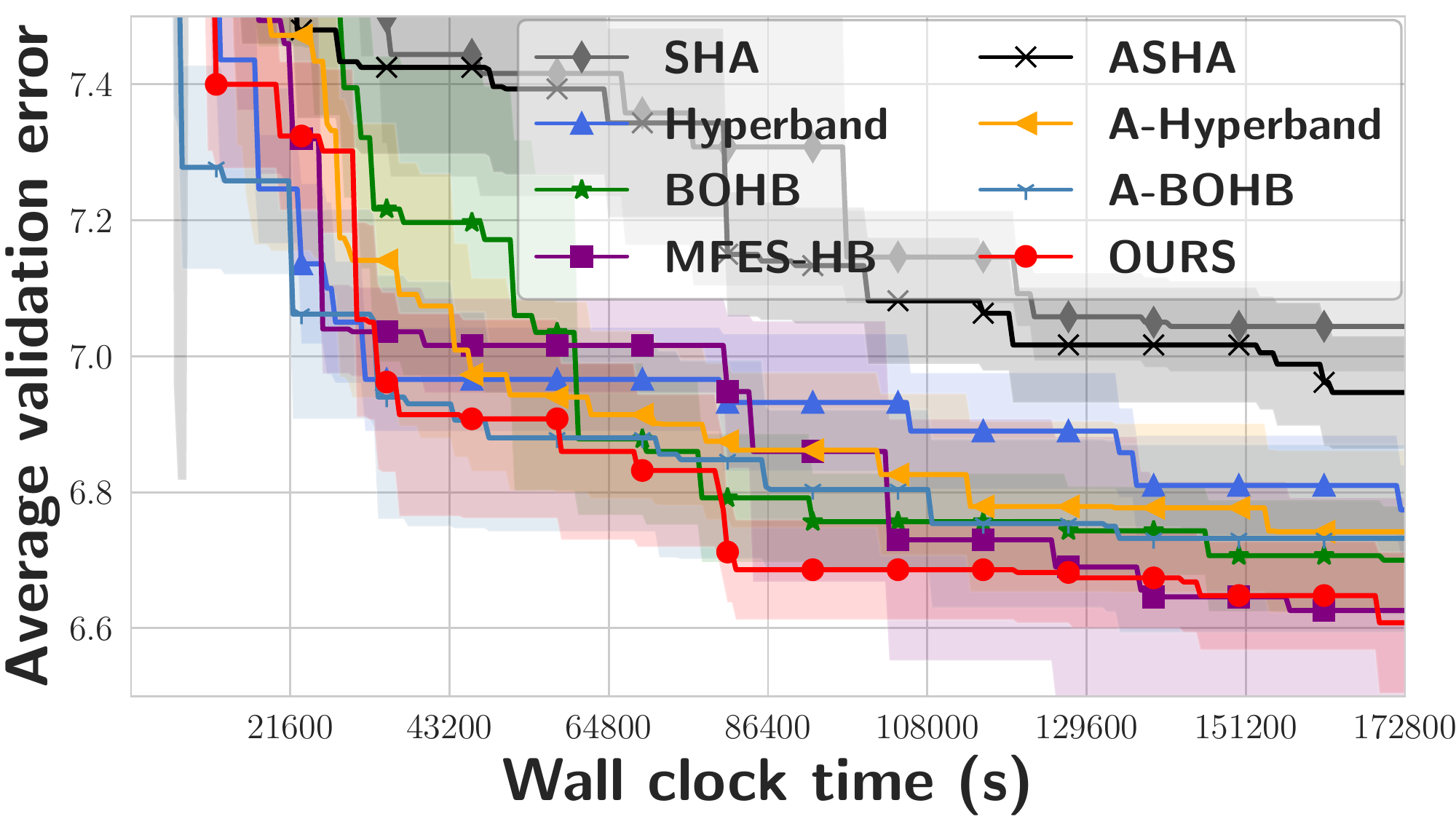}
		\label{fig:resnet}
	}}
	\vspace{-1.5em}
	\caption{Results of tuning LSTM and ResNet.}
  \label{fig:dnn}
\end{figure}

\noindent(4) \textit{Language Modelling.}
We tune a 3-layer LSTM~\cite{hochreiter1997long} on the dataset Penn Treebank. The search space includes batch size, hidden size, learning rate, weight decay and five hyper-parameters related to dropout. 
The embedding size is 400. 
The minimum and maximum number of epochs are 1 and 200. HB-based methods use 4 brackets, and the time budget is 48 hours; the default number of workers is 4, and each worker uses 8 CPU cores and 1 GPU during evaluation.

\para{Implementation Details.}
Two metrics are used in our experiments, including (1) the classification error for XGBoost tuning, ResNet tuning, and neural architecture search, and (2) the perplexity when tuning LSTM.
We use the validation and test performance stored in NAS-Bench-201 directly for neural architecture search. 
In the XGBoost tuning experiment, we randomly divide 60\% of the total dataset as the training set, 20\% as the validation set, and the left as the test set. 
In the other experiments, we split 20\% of the training dataset as the validation set. Then, we track the wall clock time (including optimization overhead and evaluation cost), and store the lowest validation performance after each evaluation. The best configurations are then applied to the test dataset to report the test performance. 
All methods are repeated ten times with different random seeds, and the mean validation performance across runs is plotted.
% As suggested in HB~\cite{li2018hyperband} and BOHB~\cite{falkner2018bohb}, $\eta$ is set to $3$ for all HB-based methods, and the probability for random sampling $\rho$ is set to 0.3. 
% For A-BOHB, we use its released version in Auto-Gluon~\cite{agtabular}.
% Furthermore, we use Pytorch~\cite{paszke2019pytorch} to train neural networks on 32 `RTX 2080Ti' GPUs, and the experiments are conducted on ten machines with 640 `AMD EPYC 7702P' CPU cores in total.
We include more experimental setups and reproduction details about \sys in the supplementary material.

\vspace{-1em}
\subsection{Architecture Search on NAS-Bench-201}
Figure~\ref{fig:nas} shows the results on NAS-Bench-201 datasets. Due to the utilization of parallel resources issue in Hyperband, asynchronous random search (A-Random) outperforms synchronous Hyperband.
% ASHA shows competitive performance in the beginning, but it requires more time to converge to the similar performance as MFES-HB in CIFAR-100 and ImageNet16-120. 
\sys obtains the best anytime and converged performance among all methods. 
Concretely, it achieves 8.2$\times$, 11.2$\times$ and 6.3$\times$ speedups against BOHB, and obtains 3.3$\times$, 2.9$\times$, and 2.0$\times$ speedups compared with A-BOHB on CIFAR-10-valid, CIFAR-100, and ImageNet16-120 respectively, which indicates its {\em superior efficiency} over the state-of-the-art methods.
In addition, \sys also gets the best test accuracy (See the results in Appendix A.5).

As reported in NAS-Bench-201~\cite{dong2019bench}, the best method is regularized evolutionary algorithm (REA)~\cite{real2019regularized}. 
For fair comparison, we also extend REA to an asynchronous version -- A-REA. 
From Figure~\ref{fig:nas}, we have that \sys shows consistent superiority over A-REA. 
Remarkably, \sys reaches the global optimum on CIFAR-100 and ImageNet-16-120 across all the ten runs, which also indicates the efficiency of \sys.

\begin{table}[tb]
\centering
\caption{Test performance on three benchmarks. (accuracy (\%) for XGBoost and ResNet, and perplexity for LSTM)}
\vspace{-1em}
\resizebox{0.99\columnwidth}{!}{
\begin{tabular}{l|cccccc}
\toprule

\multirow{2}{*}{\tabincell{c}{Method}}  & \multicolumn{4}{c}{XGBoost} & ResNet & LSTM \\
& Covertype & Pokerhand & Hepmass & Higgs & CIFAR-10 & Penn Treebank \\
\hline
{\bf Manual} & {\bf 86.91 $\pm$ 0.00} & {\bf 99.36 $\pm$ 0.00} & {\bf 87.06 $\pm$ 0.00} & {\bf 74.24 $\pm$ 0.00} & {\bf 91.88 $\pm$ 0.00} & {\bf 107.02 $\pm$ 0.00}\\ 
BO & 93.08 $\pm$ 0.19 & 99.32 $\pm$ 0.13 & 87.48 $\pm$ 0.01 & 75.40 $\pm$ 0.06 & / & / \\
SHA & 92.39 $\pm$ 0.40 & 98.43 $\pm$ 0.41 & 87.41 $\pm$ 0.02 & 75.30 $\pm$ 0.04 & 92.19 $\pm$ 0.30 & 66.05 $\pm$ 2.43 \\
Hyperband & 92.45 $\pm$ 0.37 & 98.30 $\pm$ 0.51 & 87.39 $\pm$ 0.02 & 75.29 $\pm$ 0.06 & 92.17 $\pm$ 0.31 & 65.93 $\pm$ 2.61 \\
BOHB & 92.97 $\pm$ 0.20 & 99.46 $\pm$ 0.07 & 87.44 $\pm$ 0.02 & 75.29 $\pm$ 0.05 & 92.37 $\pm$ 0.29 & 65.90 $\pm$ 2.15 \\
MFES-HB & 93.42 $\pm$ 0.08 & 99.61 $\pm$ 0.13 & 87.48 $\pm$ 0.01 & 75.44 $\pm$ 0.02 & 92.41 $\pm$ 0.24 & 64.21 $\pm$ 1.09 \\
A-Random & 91.99 $\pm$ 0.31 & 97.62 $\pm$ 0.53 & 87.38 $\pm$ 0.02 & 75.24 $\pm$ 0.07 & / & / \\
A-BO & 92.96 $\pm$ 0.09 & 99.47 $\pm$ 0.15 & 87.48 $\pm$ 0.01 & 75.35 $\pm$ 0.04 & / & / \\
ASHA & 92.61 $\pm$ 0.35 & 98.80 $\pm$ 0.24 & 87.46 $\pm$ 0.02 & 75.33 $\pm$ 0.05 & 92.23 $\pm$ 0.42 & 64.18 $\pm$ 0.58 \\
A-Hyperband & 92.39 $\pm$ 0.42 & 98.24 $\pm$ 0.46 & 87.41 $\pm$ 0.02 & 75.30 $\pm$ 0.04 & 92.16 $\pm$ 0.19 & 65.16 $\pm$ 1.22 \\
A-BOHB & 93.72 $\pm$ 0.10 & 99.83 $\pm$ 0.07 & 87.49 $\pm$ 0.01 & 75.49 $\pm$ 0.02 & 92.31 $\pm$ 0.25 & 66.02 $\pm$ 1.32 \\
\sys & \textbf{93.97 $\pm$ 0.06} & \textbf{99.93 $\pm$ 0.03} & \textbf{87.52 $\pm$ 0.02} & \textbf{75.53 $\pm$ 0.03} & \textbf{92.48 $\pm$ 0.17} & \textbf{63.54 $\pm$ 0.38} \\
\bottomrule
  \end{tabular}
  }
  \label{tab:xgb}
\end{table}

\vspace{-1.em}
\subsection{Tuning XGBoost on Large Datasets}
% In Figure~\ref{fig:xgboost} and Table~\ref{tab:xgb}, we compare \sys with ten competitive baselines by tuning XGBoost on four large datasets. 
In Figure~\ref{fig:xgboost} and Table~\ref{tab:xgb}, we compare \sys with the manual setting and ten competitive baselines by tuning XGBoost on four large datasets. 
The configurations from tuning algorithms outperform the manual settings on test results, which shows the necessity of tuning hyper-parameters for machine learning models.
Different from the other experiments, the resource type here is the subset of dataset, i.e., we use different sizes of datasets subset to perform partial evaluation if necessary. 
As BO and A-BO evaluate each configuration completely, it takes them a long time to converge to a satisfactory performance due to expensive evaluation cost (15 minutes per trial on Covertype). 
In addition, \sys and MFES-HB perform better than HyperBand, BOHB and most asynchronous methods, which indicates the advantage of leveraging low-fidelity measurements. 
Among the considered methods, \sys achieves very competitive anytime performance, and obtains the best converged performance on all of the four datasets.

\vspace{-1.em}
\subsection{Tuning LSTM and ResNet}
Figure~\ref{fig:lstm} and Table~\ref{tab:xgb} show the results of tuning LSTM on Penn Treebank. A-BOHB shows the worst converged performance among baselines, which we attribute to its failure of exploiting multi-fidelity measurements. 
A-Hyperband, MFES-HB, and \sys show similar results in the early stage (19 hours), but after that, the perplexity of A-Hyperband stops decreasing as random sampling fails to exploit history observations efficiently. 
After 150k secs (about 41 hours), \sys outperforms all baselines.

\begin{figure*}[htb]
	\centering
	\subfigure[NAS-Bench-201 on CIFAR-10-valid]{
		% Requires \usepackage{graphicx}
		\scalebox{0.23}[0.23]{
			\includegraphics[width=1\linewidth]{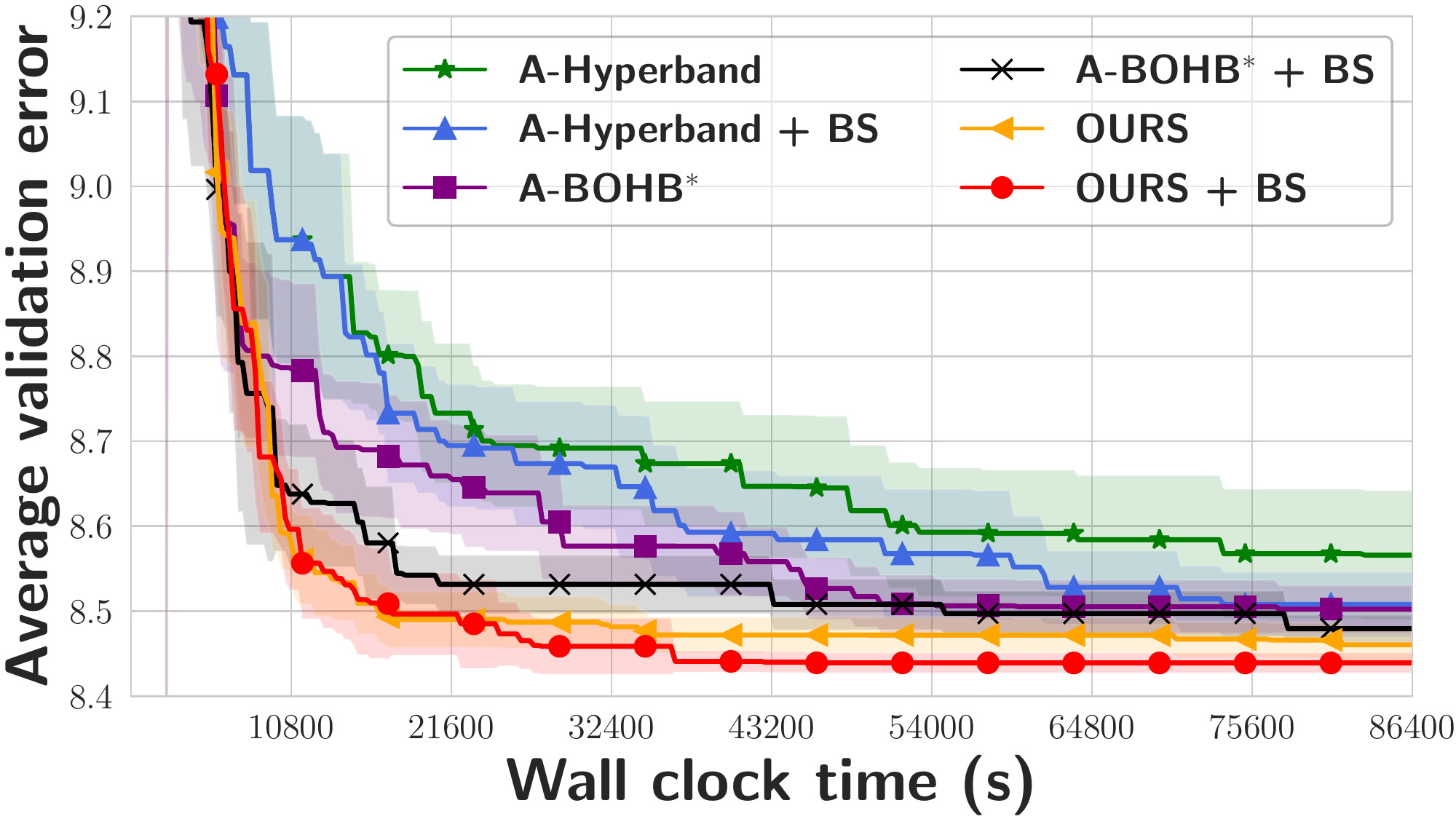}
		\label{fig:ab_cifar100}
	}}
	\subfigure[NAS-Bench-201 on ImageNet16-120]{
		% Requires \usepackage{graphicx}
		\scalebox{0.23}[0.23]{
			\includegraphics[width=1\linewidth]{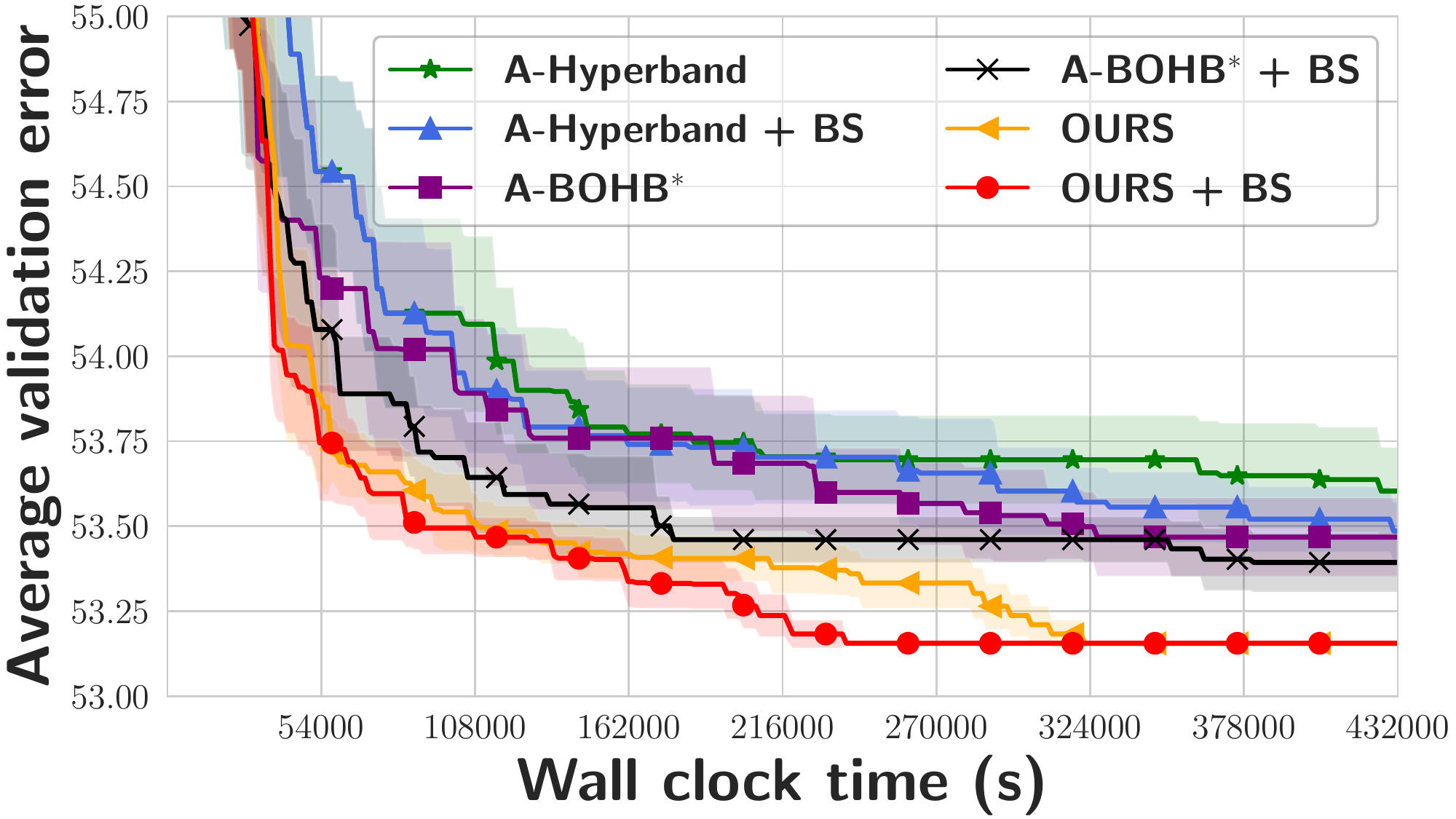}
		\label{fig:ab_imagenet}
	}}
    \subfigure[{\bf XGBoost on Covertype}]{
		% Requires \usepackage{graphicx}
		\scalebox{0.23}[0.23]{
			\includegraphics[width=1\linewidth]{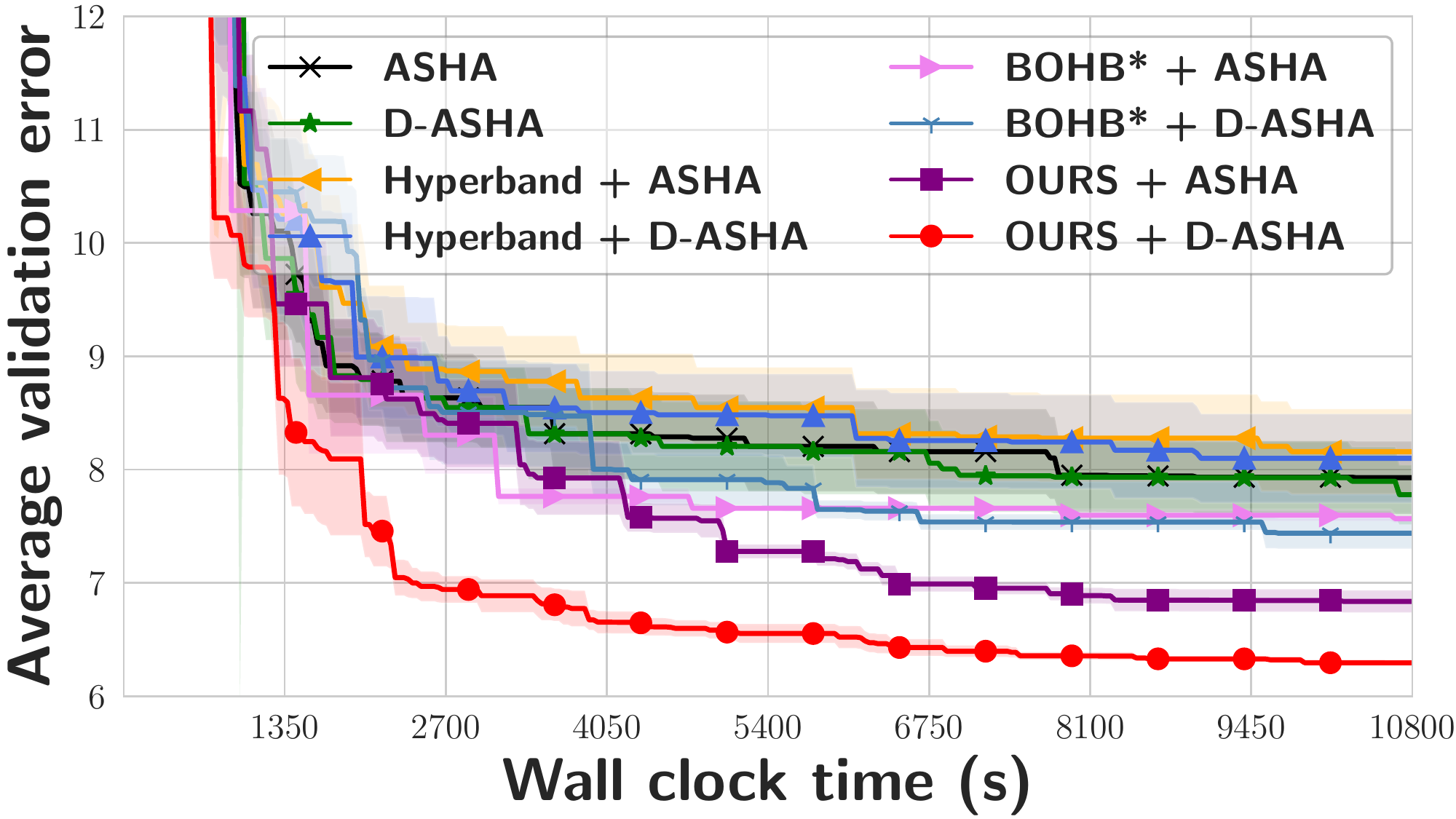}
		\label{fig:ab_covtype}
	}}
	\subfigure[{\bf XGBoost on Pokerhand}]{
		% Requires \usepackage{graphicx}
		\scalebox{0.23}[0.23]{
			\includegraphics[width=1\linewidth]{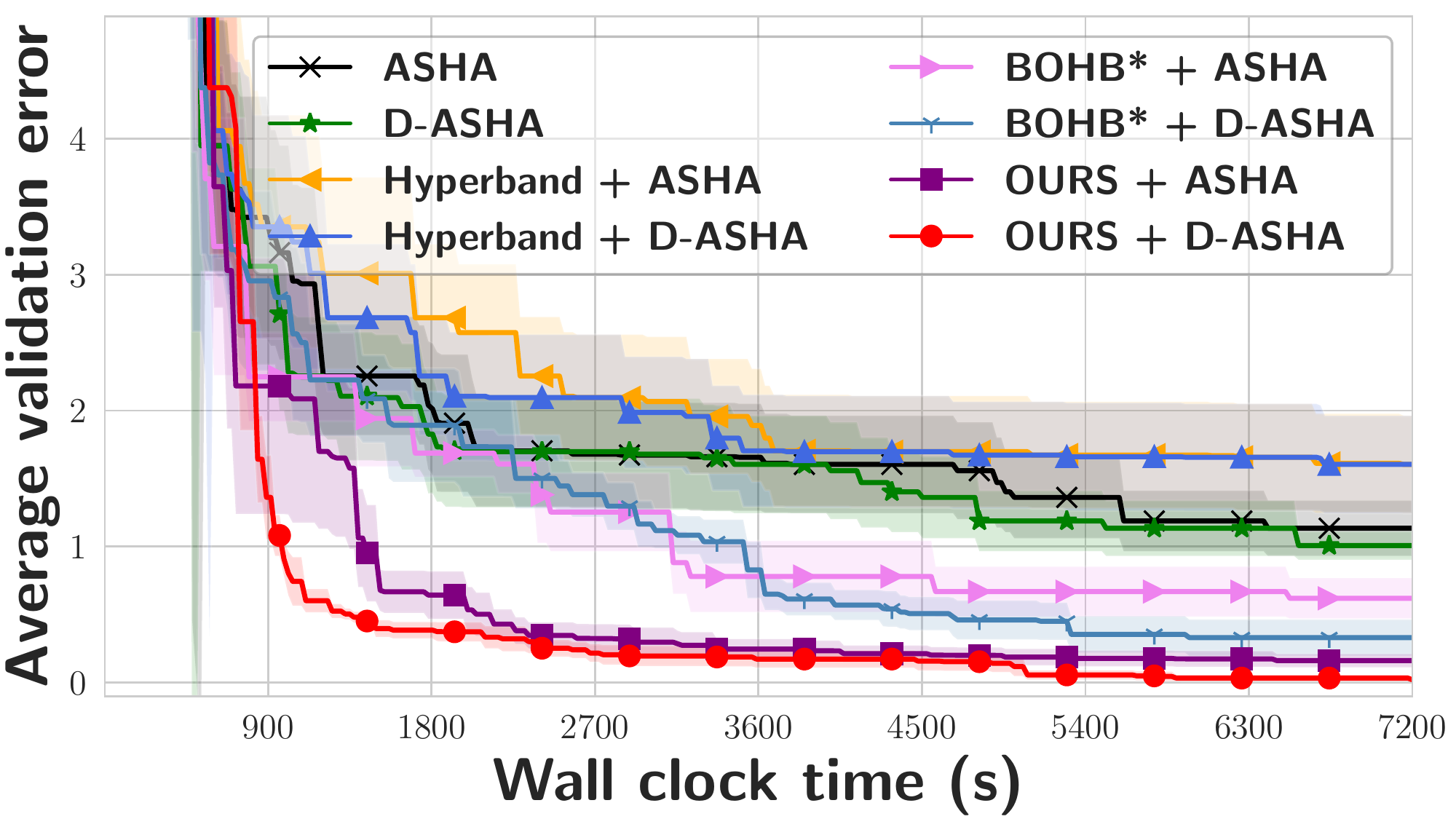}
		\label{fig:ab_pokerhand}
	}}
	\vspace{-5mm}
	\caption{Ablation studies for different components in \sys.}
	\vspace{-3.5mm}
  \label{fig:ablation}
\end{figure*}

\begin{figure}[tb]
    \vspace{-1.em}
	\centering
	\subfigure[Counting-ones Benchmark]{
		% Requires \usepackage{graphicx}
		\scalebox{0.47}[0.47]{
			\includegraphics[width=1\linewidth]{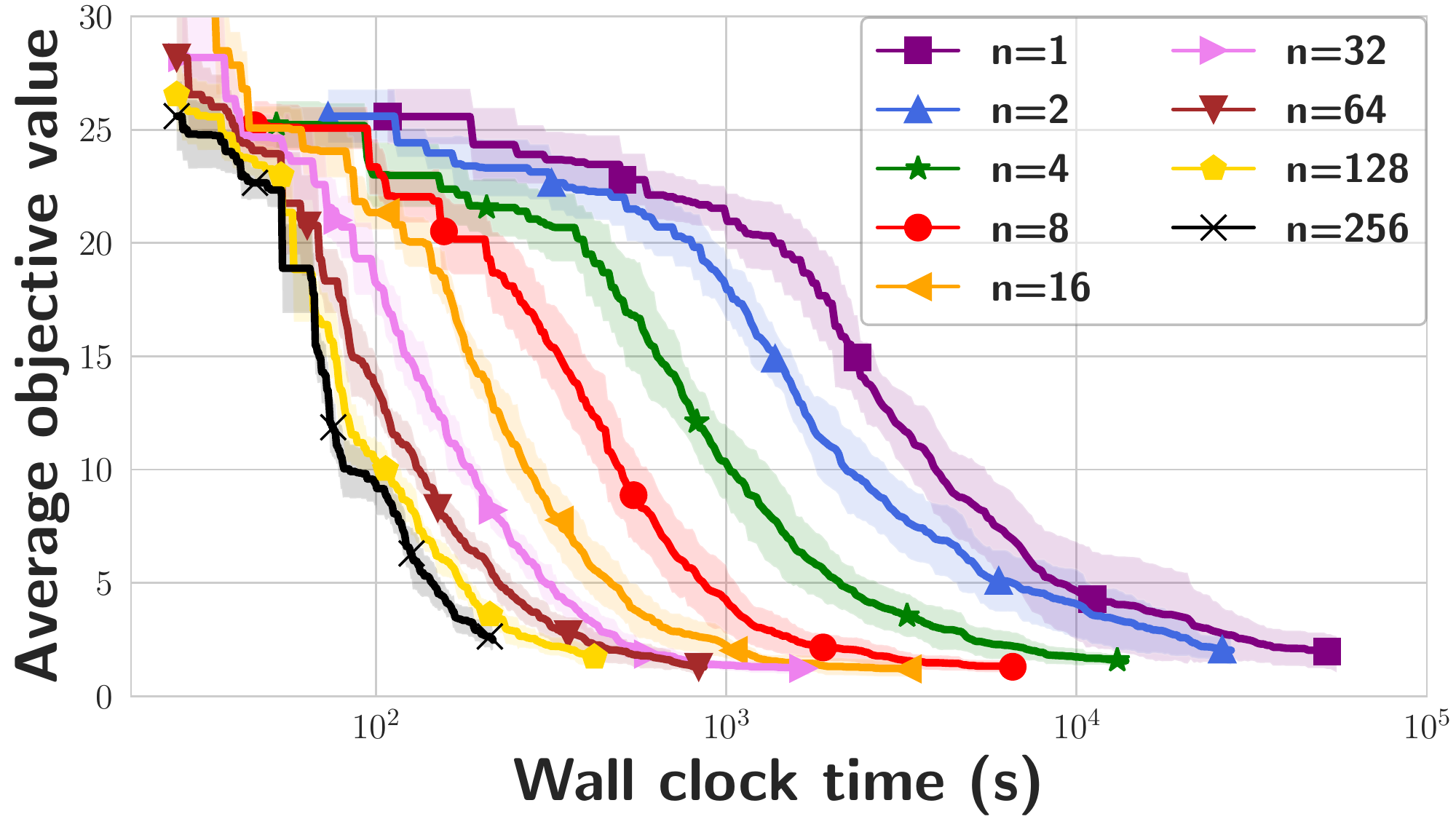}
	}}
	\subfigure[XGBoost on Covertype]{
		% Requires \usepackage{graphicx}
		\scalebox{0.47}[0.47]{
			\includegraphics[width=1\linewidth]{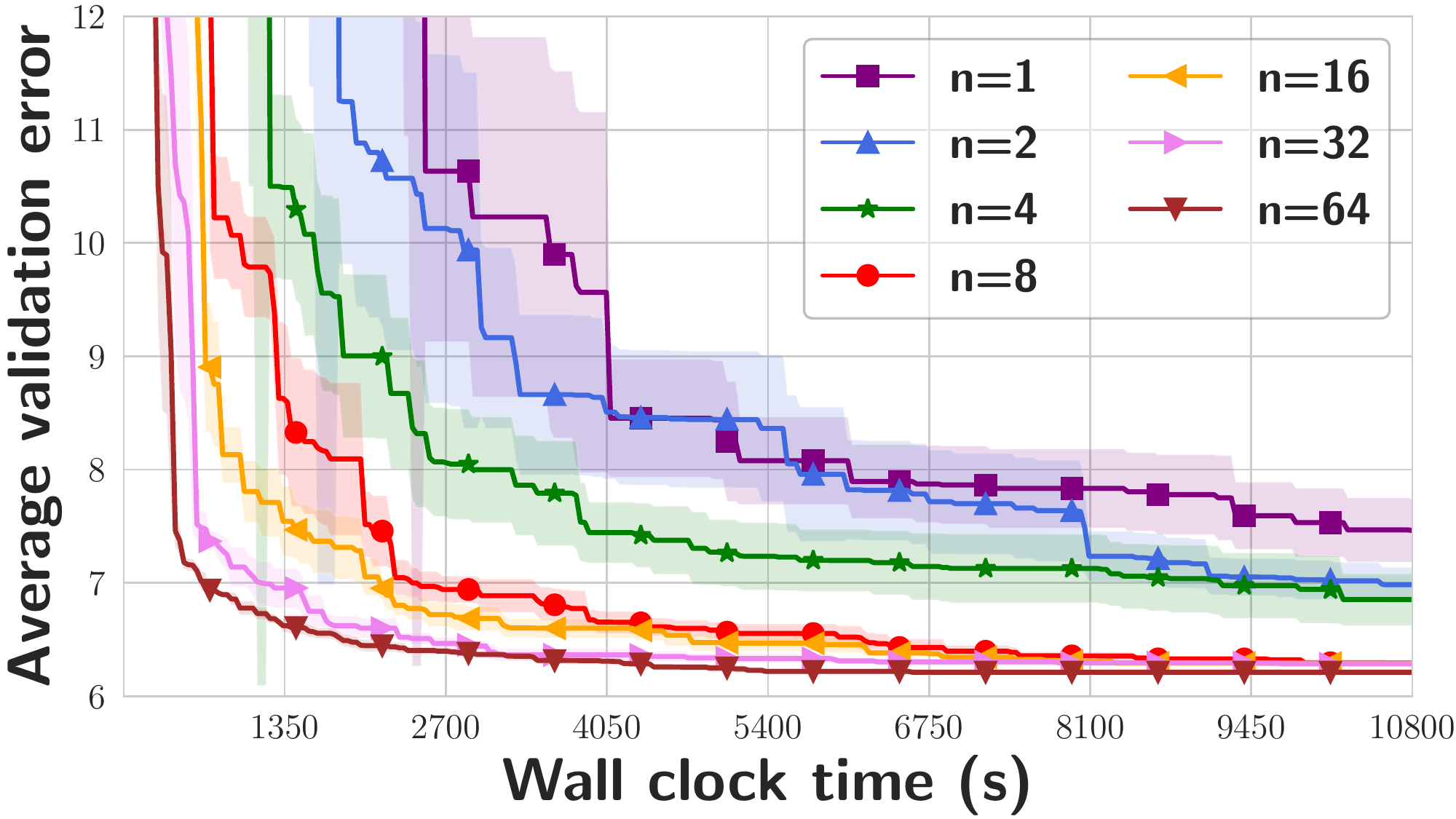}
	}}
% 	\subfigure[NAS-Bench-201 on CIFAR-100]{
% 		% Requires \usepackage{graphicx}
% 		\scalebox{0.47}[0.47]{
% 			\includegraphics[width=1\linewidth]{images/worker_nasbench201_cifar100.pdf}
% 	}}
	\vspace{-5mm}
	\caption{Scalability on the number of workers.}
  \label{fig:worker}
\end{figure}

In Figure~\ref{fig:resnet}, we display the average error of tuning ResNet on CIFAR-10. As SHA and ASHA always start evaluating each configuration from the least resources, they cannot distinguish noisy low-fidelity results, which may explain their overall worst performance. 
Though \sys and MFES-HB obtain a similar result (93.4\%), \sys shows a better anytime performance due to its asynchronous scheduling. Table~\ref{tab:xgb} shows the test result.

\vspace{-1.em}
\subsection{Scalability Analysis}
\label{sec:scale_workers}
% Previous results shows \sys works well on different scales of evaluation cost.
Figure~\ref{fig:worker} demonstrates the optimization curve with different number of parallel workers on two tuning tasks. 
We evaluate \sys by tuning the counting-ones function~
\cite{falkner2018bohb} and XGBoost on Covertype. 
The details about the counting-ones function are provided in Appendix A.4. 
To demonstrate the scalablility of \sys, we set the maximum number of workers to 256 and 64.
On both tasks, the anytime performance is better when \sys uses more workers, which indicates that \sys scales to the number of workers well. 
Notably, \sys with the maximum number of workers achieves 145.7x and 18.0x speedups compared with sequential \sys on Counting-ones Benchmark and Covertype.

\vspace{-1.em}
\subsection{Industrial-Scale Tuning Application}
In addition, we also evaluate \sys on an industrial-scale tuning task for recommendation, which aims at identifying active users. 
The dataset provided by our enterprise partner includes more than one billion instances, and we train the model using the data of seven days and evaluate it using the data of the following day.
The number of workers is 10 and the time budget is 48 hours. We evaluate ASHA, BOHB, A-BOHB and \sys, and they improve the manual setting by -0.05\%, 0.19\%, 0.35\% and 0.87\%, respectively. Moreover, we conduct an ablation study on \sys by keeping out one of the component in Table~\ref{tab:industry}.
We observe performance gain by introducing each component into \sys while Bracket Selection leads to the largest gain.
While at least one component is absent in competitive baselines, \sys improves the AUC of the second-best baseline A-BOHB by 0.54\%, which is a wide margin considering the potential commercial values.

\vspace{-1.em}
\subsection{Ablation Study}

{\em \underline{Effectiveness of Bracket Selection}. }
Figure~\ref{fig:ab_cifar100} and ~\ref{fig:ab_imagenet} illustrate the effectiveness of the proposed bracket selection method. 
We also add bracket selection (BS) to the asynchronous variant of Hyperband and BOHB. 
Note that the asynchronous BOHB here is parallelized via ASHA, but not A-BOHB mentioned in the experimental setups. 
We have that adding bracket selection helps asynchronous Hyperband, BOHB, and \sys converge better. 
In addition, in Figure~\ref{fig:ab_imagenet}, though the converged performance of \sys remains almost the same when bracket selection is employed, the anytime performance improves before 324k secs (90 hours). 
We owe this gain to the resource allocation strategy learned during optimization rather than attempting all the choices via round robin.
In this way, as few as possible training resources are automatically allocated to evaluate the configurations, and this could effectively avoid evaluating poor configurations with too many resources.

\begin{table}[t!]
\caption{Ablation study on \sys. The improvement indicates the performance gain upon manual settings.}
\vspace{-1.em}
    \centering
    \resizebox{0.7\columnwidth}{!}{
    \setlength{\tabcolsep}{5mm}{
    \begin{tabular}{lcc}
        \toprule
        Methods & Improvement (\%) &
        $\Delta$ (\%) \\
        \midrule
        w/o BS & 0.54 & -0.33\\
        w/o D-ASHA & 0.75 & -0.12\\
        w/o MFES & 0.56 & -0.31\\
        \midrule
        \sys & 0.87 & - \\
        \bottomrule
    \end{tabular}}
    }
    \label{tab:industry}
\end{table}

\noindent{\em \underline{Effectiveness of D-ASHA}. }
Figure~\ref{fig:ab_covtype} and ~\ref{fig:ab_pokerhand} show the results of applying D-ASHA to different methods. For ASHA, Hyperband and BOHB, we observe a slight improvement on both anytime and converged performance when applying D-ASHA.
For \sys, the validation error decreases by a large margin (0.5\%) on Covertype with the aid of D-ASHA.
The delay strategy could prevent the frequent promotion issue in ASHA, and further improve the sample efficiency.
Therefore, D-ASHA could achieve a higher sample efficiency while keeping the advantage of asynchronous mechanism.

\noindent{\em \underline{Effectiveness of Multi-fidelity Optimizer}. }
We compare different optimizer for configuration sampling, including random sampling (A-Hyperband + BS), high-fidelity optimizer (A-BOHB + BS), and multi-fidelity optimizer (OURS + BS). 
As shown in Figure~\ref{fig:ab_cifar100} and ~\ref{fig:ab_imagenet}, we have that surrogate-based methods outperform random sampling, while multi-fidelity optimizer outperforms high-fidelity optimizer. 
The reason is that it takes the low-fidelity measurements into consideration when selecting the next configuration to evaluate.
It also indicates that when performing hyper-parameter tuning, the low-fidelity measurements could provide useful information about the objective function, and can be used to speed up the search process.

\section{Conclusion}
\label{sec:conclusion}

In this paper, we presented \sys, an efficient and robust distributed hyper-parameter tuning framework at scale.
\sys introduces three core components targeting at addressing the challenge in the large-scale hyper-parameter tuning tasks, including (1) automatic resource allocation, (2) asynchronous scheduling, and (3) multi-fidelity optimizer.
The empirical results demonstrate that \sys shows strong robustness and scalability, and outperforms state-of-the-art methods, e.g., BOHB and A-BOHB, on a wide range of tuning tasks. 
% In the near future, we will open-source this project. 

\clearpage
\balance

%\clearpage

\bibliographystyle{ACM-Reference-Format}
\bibliography{reference}

\newpage
\appendix

\end{document}